\documentclass[preprint,12pt, authoryear]{elsarticle}

\usepackage{amsmath,amsfonts,amssymb}
\usepackage{array}
\usepackage[caption=false,font=normalsize,labelfont=sf,textfont=sf]{subfig}
\usepackage{textcomp}
\usepackage{stfloats}
\usepackage{url}
\usepackage{verbatim}
\usepackage{cite}
\usepackage{algpseudocode}
\usepackage{multirow}
\usepackage[table]{xcolor}
\usepackage{soul,pifont}
\usepackage{makecell}
\usepackage{colortbl}
\newcommand{\etal}{\textit{et al}. }
\newcommand{\ie}{\textit{i}.\textit{e}., }
\newcommand{\eg}{\textit{e}.\textit{g}., }

\begin{document}

\begin{frontmatter}

\title{Content-Aware Preserving Image Generation}

\author{Giang~H.~Le\fnref{label1}}
\ead{gianghle@seoultech.ac.kr}
\author{Anh~Q.~Nguyen\fnref{label1}}
\ead{anh.qnguyen@seoultech.ac.kr}
\author{Byeongkeun~Kang\fnref{label2}}
\ead{byeongkeun.kang@seoultech.ac.kr}
\author{Yeejin~Lee\corref{cor1}\fnref{label1}}
\ead{yeejinlee@seoultech.ac.kr}

\cortext[cor1]{Corresponding author}
\affiliation[label1]{organization={Department of Electrical and Information Engineering, Seoul National University of Science and Technology},
            addressline={232 Gongneung-ro, Nowon-gu}, 
            city={Seoul},
            postcode={01811}, 
            country={Republic of Korea}}
\affiliation[label2]{organization={Department of Electronic Engineering, Seoul National University of Science and Technology},
            addressline={232 Gongneung-ro, Nowon-gu}, 
            city={Seoul},
            postcode={01811}, 
            country={Republic of Korea}}

\begin{abstract}
Remarkable progress has been achieved in image generation with the introduction of generative models. However, precisely controlling the content in generated images remains a challenging task due to their fundamental training objective. This paper addresses this challenge by proposing a novel image generation framework explicitly designed to incorporate desired content in output images. The framework utilizes advanced encoding techniques, integrating subnetworks called content fusion and frequency encoding modules. The frequency encoding module first captures features and structures of reference images by exclusively focusing on selected frequency components. Subsequently, the content fusion module generates a content-guiding vector that encapsulates desired content features. During the image generation process, content-guiding vectors from real images are fused with projected noise vectors. This ensures the production of generated images that not only maintain consistent content from guiding images but also exhibit diverse stylistic variations. To validate the effectiveness of the proposed framework in preserving content attributes, extensive experiments are conducted on widely used benchmark datasets, including Flickr-Faces-High Quality, Animal Faces High Quality, and Large-scale Scene Understanding datasets.
\end{abstract}

\end{frontmatter}

\section{Introduction} \label{sec:introduction}
Remarkable progress has been made in computer vision tasks with the emergence of deep artificial neural networks, specifically convolutional neural networks~(CNN)~\citep{efficientnet, resnet, squeezeNET, inceptionv3, vgg, nfnet} and vision transformers~(ViT)~\citep{vit, ImproveVITbyHigh, styleSwin}. These models demonstrate remarkable performance across a spectrum of supervised vision tasks in various industrial domains,  for example, defect detection and quality control in manufacturing, anomaly detection in security applications, object recognition for robot vision~\citep{ efficientnetv2, yolov3, maskrcnn, vit, vitGANs, vitYolo, wang2022visual}, and scene understanding for autonomous vehicle~\citep{liu2021sg, lu2023transflow, han2024prototypical}.
However, the process of data cleaning and annotation required for supervised tasks are highly costly~\citep{imagenet, coco, cifar10}, making it challenging to apply these networks without readily available annotations. 

To address this, generative models have emerged as promising solutions. For instance, generative adversarial networks~(GANs) allow for high-quality image generation that closely aligns with the desired dataset distribution, offering a way to obtain the data needed for vision tasks~\citep{msggans, styleganADA, stylegans, stylegans2, acGANs, stargans, starganv2, alias-free, zhou2021survey, mahendren2023diverse}. Similarly, recent diffusion models~\citep{ho2020denoising, song2020denoising, zhuang2023diffusion, wang2023diffusion} have gained attention for their ability to produce high-quality images. With advancements in large language models, diffusion models have increasingly focused on integrating visual models with other modalities, such as text encoders, to enable text-conditional generation. This integration makes them particularly effective for tasks like text-to-image synthesis~\citep{rombach2022high, hertz2022prompt, han2024image}. While diffusion models are currently preferred for text-based applications, GANs still have significant potential, especially in scenarios focused solely on image-based applications. Once trained, GANs offer the advantage of faster image generation and have been very successful in producing sharp, high-quality images where their specific strengths can be fully exploited. 

\begin{figure}[t]
  \centering
  \begin{tabular}{@{}c@{~}c@{~}}
    \includegraphics[width = 0.37\textwidth]{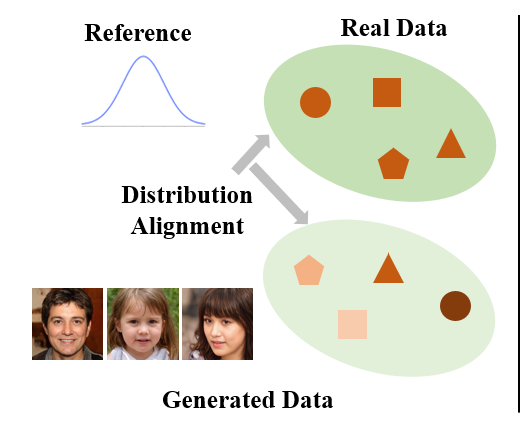} \hspace{0.2cm} &
    \includegraphics[width = 0.40\textwidth]{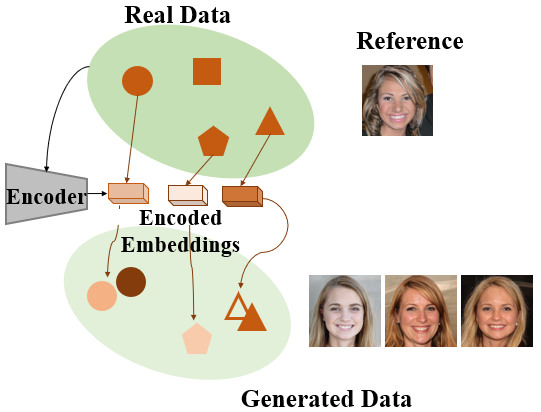} \\
    {\footnotesize (a)} &
    {\footnotesize (b)}   
  \end{tabular}
  \caption{Comparison of (a) typical image generation with (b) the proposed content-preserving image generation. Typical GANs generate random images following the distribution of real images. In contrast, the proposed framework allows users to exert control over image generation, enabling them to specify desired content attributes in the generated images.} \label{fig:concept}
\end{figure}

The GAN is a framework that estimates generative models through an adversarial process~\citep{origingans}. GANs have played a significant role in various computer vision applications and have been an active research area since their invention. Despite the considerable success of GANs in the literature, effectively managing and controlling specific contents of generated images--such as underlying spatial structure or precise attributes--remains a challenging task. This challenge stems from the fundamental training objective of GANs, which primarily focuses on mapping the distribution of output images to an input distribution rather than explicitly generating images with desired contents~\citep{origingans, stylegans,stylegans2,msggans}, as depicted in Figure~\ref{fig:concept}(a). Consequently, generating images that meet users' requirements using GANs is still a difficult problem. 

To overcome such a challenge, several methods have been proposed to impose specific constraints on the generated images, aiming to gain control over the image generation process. One approach is to convert GANs into a supervised learning manner by integrating auxiliary information, such as class labels or reference images, to guide the content of the generated outputs~\citep{cGANs}. This is done by concatenating additional inputs with random noise and feeding them into the networks. Additionally, researchers have proposed that discriminators can perform the classification task using categorical labels in conjunction with the critic to distinguish between real and fake samples. This is usually achieved by using a cross-entropy loss in the objective function~\citep{acGANs}. This approach has shown the possibility of controlling the content of generated images~\citep{stargans, starganv2}. However, manually annotating training images is highly labor-intensive and time-consuming. Moreover, widely-used benchmark datasets for image generation, such as Flickr-Faces High Quality~(FFHQ)~\citep{stylegans}, Animal Faces High Quality~(AFHQ)~\citep{stargans, starganv2}, and Large-scale Scene Understanding~(LSUN)~\citep{lsun}, are typically unlabeled. Hence, converting unsupervised tasks into supervised learning models may not be a practical solution to address the challenge of efficiently controlling the content of generated images by GANs.

Another recent approach involves understanding the latent space of GANs and using it to control image generation~\citep{latentSpace, stylegans}. These methods typically utilize a multilayer perceptron encoder to transform the Gaussian noise input into a disentangled distribution of the same dimension. By using low-distance samples from this disentangled distribution, it becomes easier to generate images with the desired style. However, these methods do not have the ability to directly generate feature vectors from real-world images that would influence the output images. Instead, they only provide control over output images through a predefined set of synthesized samples. More recently, an unsupervised method was proposed in~\citep{redGANs} to control the attributes of generated images by finding the optimal direction in the latent space based on the differential changes in different feature sets. However, the acquisition of the feature sets requires additional models, such as a face recognition model for identity features or an age regression model for age features.

To tackle the previously mentioned labeling challenge and contending with constraints on control in limited settings, we propose a novel framework that can generate high-fidelity images with specific desired content characteristics as requested by users. The proposed framework leverages a real-world image to guide the generation process, resulting in output images that closely resemble the content of the input image, as illustrated in Figure~\ref{fig:concept}(b). To achieve this, we introduce an encoding module incorporating two essential components: an encoder module for extracting content features and a content fusion module for generating refined content-guided vectors. This module analyzes generated images with known feature embeddings to determine the direction and intensity of content perturbations. Using this analyzed content feature vector, the proposed framework can produce images that preserve the contents of the guided images without being confined to predefined sets or constraints. 
Specifically, the encoding module is designed to concentrate solely on selected frequency components linked to intriguing content features. Its aim is to adeptly learn desired features while excluding undesired features from the learning process. The effectiveness of frequency selection in the generation process is verified by conducting an analysis to explore the impact of frequency bands on the overall process. This analysis provides convincing evidence that frequency analysis can serve as a meaningful guide in determining which components should be processed effectively to enhance content preservation and image quality. This finding also aligns with recent evidence that manipulating the frequency components of input images can significantly improve model performance~\citep{imageConv2FreqConv,yin2019fourier,schwarz2021frequency, wang2020high}. 

In summary, this work makes the following contributions: 
\begin{itemize}
    \item We propose a novel framework, named Content-Aware Preserving Image Generation Framework~(CAP-GAN), which enables image generation with explicit control over the content of the output images, ensuring content fidelity with guiding images.
    \item We investigate the impact of frequency bands on image generation and demonstrate that frequency analysis can serve as a valuable guide for content preservation and image quality improvement.
    \item Extensive experiments have been conducted to validate the effectiveness of the proposed framework in accurately predicting the content of the generated images. 
\end{itemize}

The remainder of this paper is organized as follows: Section~\ref{sec:related_works} provides a comprehensive review of various types of GANs and diffusion models, and it explores the role of frequency analysis in vision processing. In Section~\ref{sec:proposed_method}, we present a detailed description of the proposed framework, including the structure of each block and the underlying design principles. We also explain the process of extracting frequency components and their seamless integration into the proposed framework. 
Section~\ref{sec:results} presents the experimental results and additional studies, offering a thorough analysis and interpretation of the findings. Finally, in Section~\ref{sec:conclusion}, we conclude the paper with final remarks. 

\section{Related Work} \label{sec:related_works}
\subsection{Content Control in Generative Models}
\label{sec:gan_content_control_review}

Generative models have proven successful in image generation tasks. However, predicting the precise content of the outputs remains challenging due to the inherently stochastic nature of the generation process. To address this issue, some research has introduced methods that enable control over the details of the generated outputs. 

One approach uses annotated information, such as labels, to improve control over the generated outputs. For example, the auxiliary classifier GAN (AC-GAN) incorporates the embedded label into the input latent noise~\citep{acGANs}. In AC-GAN, the discriminator plays a dual role: it distinguishes between real and fake samples and predicts the expected label of the input image. Other methods~\citep{stargans, starganv2} have also attempted to control the attributes of the generated images using multi-class labels. Additionally, early research on diffusion models~\citep{dhariwal2021diffusion} has shown that, with classifier guidance, these models can achieve sample quality comparable to that of GANs. To eliminate the need for classifiers, the work~\citep{ho2022classifier} proposes a method that jointly trains a conditional and an unconditional diffusion model and combines their results. However, this approach still requires annotated labels during training.  

Another approach is to extract latent codes to control the image generation process. For example, MixNMatch~\citep{mixnmatch} generates latent vectors using four different encoders and combines these vectors to control different aspects of the generated content, with each encoder specializing in features such as background, object pose, shape, and texture. Similarly, the work~\citep{redGANs} manages image attributes by finding the optimal direction in latent space based on differential changes between feature sets, using an auxiliary face identification model and a landmark extraction model. HistoGAN~\citep{histogan} controls color information in generated images by analyzing differentiable histograms of colors in log-chroma space. The diffusion-GAN model~\citep{wang2023diffusion} integrates a diffusion model into the GAN framework, using the diffusion model as a generative component while applying adversarial training techniques to improve sample quality. The diffusion probabilistic field~(DPF)~\citep{zhuang2023diffusion} also captures latent information by parameterizing it as additional variables.

Note that the methods mentioned above, whether GAN-based or diffusion-based, can generate high-quality images but often require annotated information, such as labels. Even those methods that do not rely on annotations typically lack explicit mechanisms for controlling the content of the generated images. In contrast, the proposed framework offers significant advantages. It allows for explicit control over the content of output images by using latent embeddings derived from real-world images, focusing solely on the desired content while suppressing undesired components. This approach eliminates the need for additional annotation information.
Moreover, the proposed framework enables content-preserving image generation with only a single generator, encoder, and discriminator, avoiding the complexity of using multiple encoders, generators, and discriminators. This streamlined architecture not only simplifies the model but also reduces computational cost and resource requirements, making it a more efficient and practical solution. 

\subsection{Frequency Analysis in Neural Networks} 
Previous research has extensively explored the relationship between frequency components and image perception, emphasizing the significance of both high- and low-frequency information~\citep{oliva2006hybrid, petras2019coarse, lee2022lossless}. Early work in \citep{oliva2006hybrid} conducted experiments with combined images of two to demonstrate that high-frequency components are crucial for object recognition, capturing fine details and edges, and low-frequency components contribute to the perception of global scene layout and semantic context.  Additionally, Petras \etal observed that low spatial frequencies provide coarse information that guides the integration of finer and high spatial frequency details, contributing to a more comprehensive understanding of visual input~\citep{petras2019coarse}. More recently, the work in \citep{lee2022lossless} proposed a lossless image compression method that considers both low- and high-frequency components. This study highlighted the importance of preserving global structural information and fine details simultaneously, leading to improved compression outcomes.

Building upon the aforementioned findings, researchers have investigated the use of frequency information in deep neural networks for various computer vision tasks~\citep{convFreq, imageConv2FreqConv,yin2019fourier}.
Likewise, in GANs, frequency-based methods have been explored for generating high-quality images~\citep{schwarz2021frequency, wang2020high, waveletGAN, highFreqOnGANs, ssdGAN, watchUU, dzanic2020fourier, fuoli2021fourier, khayatkhoei2022spatial, freqAnalysisGAN}. SWAGAN~\citep{waveletGAN} is an example of using wavelet transformations to train models in the frequency domain, producing better results for small details than traditional methods. Some works have explored incorporating the frequency spectrum into the discriminator module~\citep{highFreqOnGANs, ssdGAN} or adding it as a regularization term to the objective function~\citep{watchUU}. These works aim to enhance the generation of synthetic images by emphasizing the frequency details. Other studies have highlighted the disparity between real and synthetic images in the frequency domain, despite their visual similarity in the spatial domain. To address this issue, some works focus on utilizing augmentation techniques to bridge the gap in the frequency domain~\citep{dzanic2020fourier}. Others utilize frequency information directly in the training process~\citep{fuoli2021fourier} or manipulate the upsampling methods within the generator module~\citep{khayatkhoei2022spatial, freqAnalysisGAN}. 

The previously mentioned studies have shown that converting images in the spatial domain to the frequency domain can help models capture important features that might be difficult to extract in the image domain. 
However, this approach also carries a risk of losing features that are only present in the image domain. As such, the proposed work develops a different strategy. We initially process the image in the frequency domain, implementing techniques to eliminate redundant features and extract content information. Subsequently, we convert this processed information back to the image domain, allowing our model to leverage additional information in the spatial domain. This approach retains both the advantages of frequency-based methods in capturing trimmed meaningful features and the benefits of working directly in the spatial domain to preserve unique structural features.

\section{Proposed Methodology}\label{sec:proposed_method}

Considering the challenges of controlling the content of generated images solely from a random noise vector, we propose a framework that incorporates real-world images to guide the encoding of desired content features. In Section~\ref{sec:architecture}, we present the overall structure of the proposed framework, outlining the main components and their interactions. In Section~\ref{sec:proposed_encoder}, we outline the training strategy for each module in the framework and explain the rationale behind our design choices.

\subsection{Overview}\label{sec:architecture}
\begin{figure*}[t]
\begin{minipage}[c]{0.8\textwidth}
    \centering
    \includegraphics[width=0.95\textwidth]{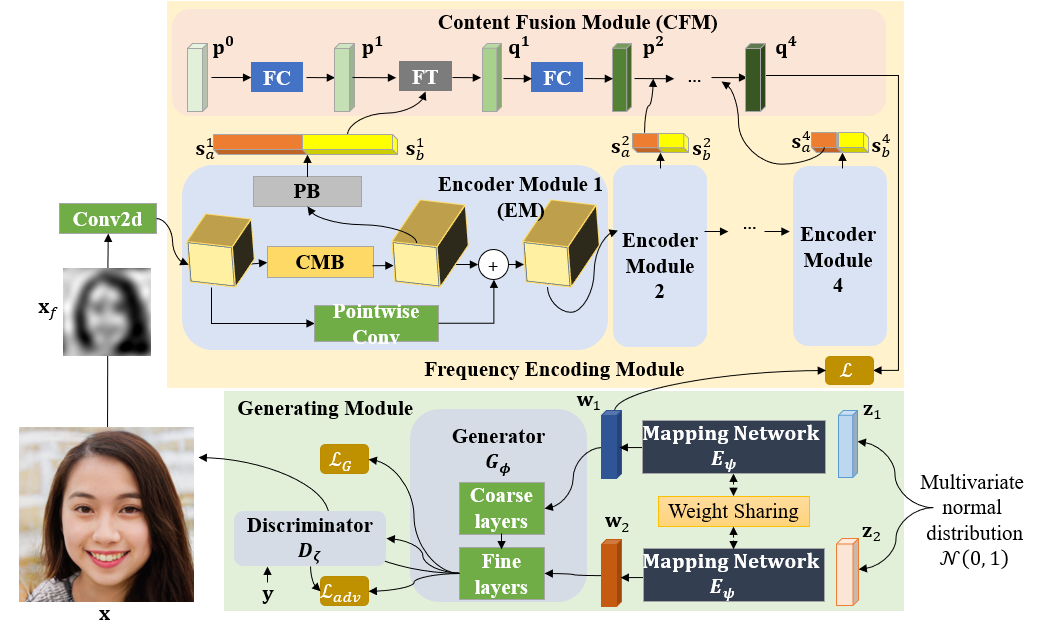} \vspace{-0.2cm}\\
    {\footnotesize (a)}
\end{minipage}
\vspace{0.1cm}
\begin{minipage}[c]{0.15\textwidth}
    \centering
    \includegraphics[width=0.86\textwidth]{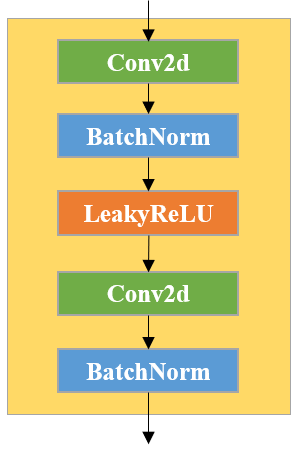}  \\
    {\footnotesize (b) CMB} \vspace{0.2cm}\\
    \includegraphics[width=0.86\textwidth]{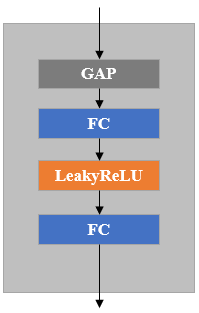}  \\
    {\footnotesize (c) PB} \\
\end{minipage} \vspace{-0.1cm}
\caption{Description of main components of the proposed framework during training. (a) Overall structure. (b) Composition of component manipulation block~(CMB). (c) Composition of projecting block~(PB). During the training process, the frequency encoding module generates a content-guiding vector from a frequency-analyzed image. This vector is designed to closely align with the feature embedding of the reference image $\mathbf{x}$. This alignment enables effective control over the content of the generated images during the inference stage. Note that BatchNorm in CMB refers to batch normalization~\citep{batchnorm}, GAP in PB stands for global average pooling, FC denotes a fully connected layer, and FT represents feature transformation.} \label{fig:overall_structure}
\end{figure*}

\begin{figure}
    \centering
    \includegraphics[width=0.85\textwidth]{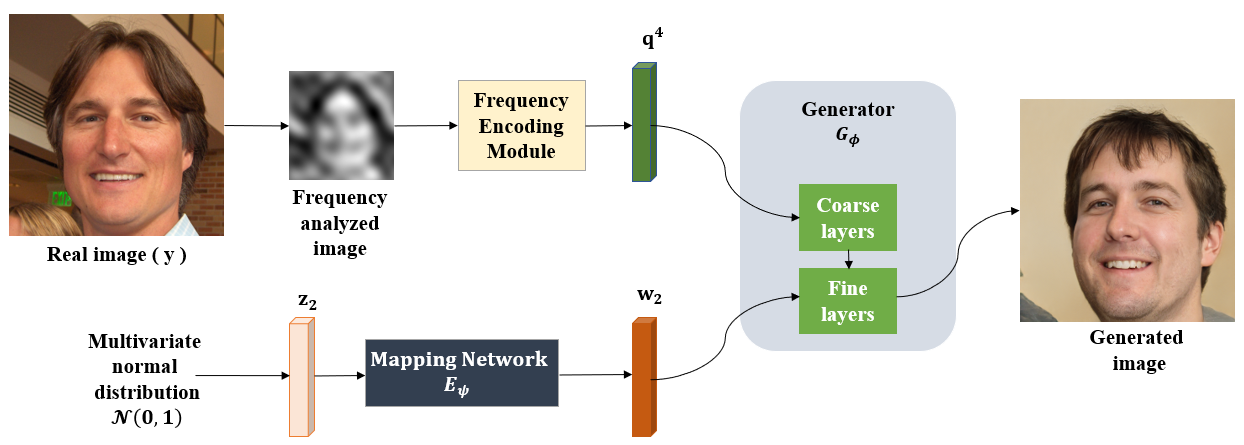} \\ 
    \caption{Inference phase of the proposed framework. The trained frequency encoding module takes a real image as an input, extracts content features from it, and transfers these features to the generator. \vspace{-0.2cm}}\label{fig:inference}
\end{figure}

Figure~\ref{fig:overall_structure} provides an overview of the architectural design of the proposed framework during training, and Figure~\ref{fig:inference} illustrates the process for inference. The proposed framework consists of two main modules: a generating module and a frequency encoding module. 

\noindent \textbf{Generating Module.} This module aims to synthesize images from random vectors and includes three components: the mapping network $E_{\mathbf{\psi}}$, the discriminator $D_{\mathbf{\zeta}}$, and the generator $G_{\mathbf{\phi}}$, which are adopted from StyleGAN2. 

The generator $G_{\mathbf{\phi}}$, parameterized by $\mathbf{\phi}$, takes the vectors $\mathbf{w}_{\scriptscriptstyle 1} \in \mathbb{R}^{512}$ and $\mathbf{w}_{\scriptscriptstyle 2} \in \mathbb{R}^{512}$ as inputs and produces synthesized images $\mathbf{x} \in \mathbb{R}^{256\times256\times3}$. The vectors  $\mathbf{w}_{\scriptscriptstyle 1}$ and $\mathbf{w}_{\scriptscriptstyle 2}$ are projected by the mapping network $E_{\mathbf{\psi}}$, parameterized by $\psi$, from $\mathbf{z}_{\scriptscriptstyle 1}$ and $\mathbf{z}_{\scriptscriptstyle 2}$. The vectors $\mathbf{z}_{\scriptscriptstyle 1}$ and $\mathbf{z}_{\scriptscriptstyle 2}$ are randomly sampled from a standard normal distribution $\mathcal{N}(0,1)$. Once $G_{\mathbf{\phi}}$ produces images $\mathbf{x}$, the discriminator $D_{\mathbf{\zeta}}$ evaluates them, along with real-world image $\mathbf{y} \in \mathbb{R}^{256\times256\times3}$ to determine whether they should be classified as real or fake.

\noindent \textbf{Frequency Encoding Module.} This module aims to encode the final content-guiding vector $\mathbf{q}^4 \in \mathbb{R}^{1024}$ from the frequency-refined image $\mathbf{x}_f$, as shown in Figue~\ref{fig:overall_structure}. The vector $\mathbf{q}^4 \in \mathbb{R}^{512}$ is then used as an input to $G_{\mathbf{\phi}}$, replacing $\mathbf{w}_{\scriptscriptstyle 1}$, to control the content of generated images at inference.

This module consists of two components: the content fusion module~(CFM) and the encoder module~(EM). The role of the EM is to progressively extract features from an input image. This process begins by taking a feature map derived from a frequency-refined intensity image $\mathbf{x}_{\scriptscriptstyle f} \in \mathbb{R}^{128\times128}$ and then producing the intermediate content-guiding vectors $\mathbf{s} \in \mathbb{R}^{1024}$ through the component manipulation block~(CMB) and the projecting block~(PB). 
The role of the CFM is to train learnable constant vectors and integrate them with $\mathbf{s}$ from the EM blocks. Specifically, the intermediate content-guiding vectors $\mathbf{s}$ are combined with the vectors $\mathbf{p} \in \mathbb{R}^{512}$, trained through the CFM, and then transformed into the final content-guiding vector $\mathbf{q}^4 \in \mathbb{R}^{512}$. 

\noindent \textbf{Training Strategy.} 
It is well established that the effects of injecting styles into style-based generators depend on the layers to which the styles are applied~\citep{stylegans,stylegans2, bigGANs, styleganADA}. Style vectors injected into coarse and middle~(lower) layers~(denoted as $\mathbf{w}_1$ in Figure~\ref{fig:overall_structure}) mainly affect the high-level features of an image, such as pose, general hairstyle, or face shape. In contrast, style vectors injected into finer~(higher) layers~(denoted as $\mathbf{w}_2$ in Figure~\ref{fig:overall_structure}) play a key role in defining background details or color information. 
Based on this understanding, the proposed method is designed to control image content by injecting the desired content styles into the coarse and middle layers, which are produced by the proposed frequency encoding module. Specifically, as shown in Figure~\ref{fig:inference}, during inference, the generator $G_{\mathbf{\phi}}$ produces images using two vectors: a content-guiding vector $\mathbf{q}^4$ extracted from a real-world image and a noise-projected vector $\mathbf{w}_{\scriptscriptstyle 2}$. The content-guiding vector $\mathbf{q}^4$, obtained from the frequency encoding module, is injected into the coarse~(lower) layers of the generator. Meanwhile, the vector $\mathbf{w}_{\scriptscriptstyle 2}$, projected from a noise vector, is injected into the fine~(higher) layers to introduce variation to the output images. 

\subsection{Training} \label{sec:proposed_encoder}
To achieve the training goal, the proposed framework is trained in a two-phase process: 1) in the first phase, the generation module is optimized to generate images based on the projected noise vectors, and 2) in the second phase, the frequency encoding module is trained to generate content-guiding vectors, while keeping the parameters of the generation module frozen.

\subsubsection{Phase I: Training Generating Module}\label{sec:generating_module}
This training phase has two primary objectives: to optimize $G_{\mathbf{\phi}}$ and $E_{\mathbf{\psi}}$ to generate high-fidelity images that closely match real-world image distributions, and to optimize $D_{\mathbf{\zeta}}$ to accurately distinguish the generated images as fake. 
The generator $G_{\mathbf{\phi}}$ is optimized using the loss determined by the evaluation of generated images: 
\begin{align}
    \mathcal{L}_{\scriptscriptstyle G} = - \frac{1}{N_b}\sum\limits^{N_b}_{j=1} \log \left(\sigma \left(D_{\scriptscriptstyle \zeta}\left( G_{\scriptscriptstyle \phi} \left(  \mathbf{w}_{\scriptscriptstyle 1, j}\right) \right) \right) \right), 
\end{align}
where $N_b$ is the number of images in a mini-batch, $j$ denotes the image index, and $\sigma\left( \cdot \right)$ denotes the sigmoid function.
Subsequently, the discriminator of this module is trained using the adversarial loss with a regularizer of the squared norm of the gradients of the discriminator on real images~\citep{r1reg}:
\begin{align}\begin{array}{l}
    \mathcal{L}_{adv}\!=\!\frac{1}{N_b}\sum\limits^{N_b}_{j=1}  \Big[ \log  \left( \sigma \left( D_{\scriptscriptstyle \zeta}\left(\mathbf{y}_{\scriptscriptstyle j}\right)\right)\right) + \log \left( \sigma \left( 1-D_{\scriptscriptstyle \zeta} \left(G_{\scriptscriptstyle \phi} \left(  \mathbf{w}_{\scriptscriptstyle 1, j}\right) \right)\right)\right) \Big. \vspace{0.1cm}\\ 
    \hspace{2.6cm} +\: \Big. \lambda \lvert\lvert \nabla D_{\scriptscriptstyle \zeta}\left(\mathbf{y}_{\scriptscriptstyle j} \right) \rvert\rvert^{\scriptscriptstyle 2}_{\scriptscriptstyle 2} \Big], 
\end{array} \end{align}
where $\lambda$ is a weighting factor to control the influence of each loss on the total loss function, $\nabla$ denotes the gradient of a function, and $\lvert \lvert  \cdot \rvert \rvert_{\scriptscriptstyle 2}$ denotes $\ell2$-norm. 

\subsubsection{Phase II: Training Frequency Encoding Module}\label{sec:frequency_encoding_module}
The goal of the second training phase is to optimize the parameters of all components in the frequency encoding module. This is achieved by utilizing frequency-refined images that emphasize high-level content features while suppressing undesired fine details. The procedure for refining these frequency components is outlined in the section ``Frequency Selection.'' Using these refined inputs, the encoder modules analyze the patterns to produce intermediate content guide vectors~(as detailed in the section of ``Guiding Vector Generation''). These intermediate vectors are then used to produce the final content guide vectors~(
See the ``Content Fusion'' section).
This process ensures that the extracted features align closely with the embeddings of reference images. The loss function used to measure this alignment is discussed in the ``Loss Design'' section.

\noindent \textbf{Guiding Vector Generation. } 
In the frequency encoding module, four EMs are connected in sequence ($i = 1, 2, 3, 4$) to extract fine features from $\mathbf{x}_{\scriptscriptstyle f}$ and produce distilled guiding vectors ($\mathbf{s}^{\scriptscriptstyle 1}, \cdots, \mathbf{s}^{\scriptscriptstyle 4}$). These distilled vectors are then used to generate the final content-guiding vector $\mathbf{q}^4$. In each EM, feature maps are processed by the CMB for further feature extraction. The resultant feature maps are then combined with the input feature maps using a $1\times1$ convolution. This addition enhances the depth of the blocks by leveraging identity mappings and residual connections, which helps mitigate training issues such as exploding or vanishing gradients during backpropagation~\citep{resnet}. 
Meanwhile, the feature maps produced by the CMB are passed through the PB, which transforms them into the distilled guiding vector,  $\mathbf{s}^{\scriptscriptstyle i} = \left[ \mathbf{s}_{\scriptscriptstyle a}^{\scriptscriptstyle i},\:\: \mathbf{s}_{\scriptscriptstyle b}^{\scriptscriptstyle i}\right] \in \mathbb{R}^{1024}$ for the $i$-th EM. These distilled vectors are divided into two parts: the upper part $\mathbf{s}_{\scriptscriptstyle a}^{\scriptscriptstyle i} \in \mathbb{R}^{512}$ is responsible for modifying the direction of contents based on the extracted features while the lower part $\mathbf{s}_{\scriptscriptstyle b}^{\scriptscriptstyle i} \in \mathbb{R}^{512}$ takes responsible for translating the modified vector.

\noindent \textbf{Content Fusion.} The content fusion module comprises four fully connected~(FC) layers followed by a feature transform (FT) layer. 
This module initiates with a learnable constant vector $\mathbf{p}^{\scriptscriptstyle 0}$, which is initialized by sampling a randomly generated vector from a normal distribution with zero mean and unit variance, $\mathcal{N}(0,1)$. The initial vector $\mathbf{p}^{\scriptscriptstyle 0} \in \mathbb{R}^{512}$ is then processed with the guiding vectors to incorporate its information. Concretely, followed by each fully connected layer, the activation $\mathbf{p}^{\scriptscriptstyle i}$ is transformed into the modified feature embedding $\mathbf{q}^{\scriptscriptstyle i}$ through scaling and translation with the content-guiding vector, as follows:
\begin{align} \label{eq:adain}
    \mathbf{q}^{\scriptscriptstyle i} = 
    \mathbf{s}_{\scriptscriptstyle a}^{\scriptscriptstyle i} \odot \mathbf{p}^{\scriptscriptstyle i} + 
    \mathbf{s}_{\scriptscriptstyle b}^{\scriptscriptstyle i},
\end{align}
where $\odot$ denotes the Hadamard product, and $i$ denotes the block index ranging from $1$ to $4$. In (\ref{eq:adain}), $\mathbf{s}^{\scriptscriptstyle i}_{\scriptscriptstyle a}$ controls the direction of the content-guiding vector, and $\mathbf{s}^{\scriptscriptstyle i}_{\scriptscriptstyle b}$ controls intensity of the content perturbations. 

\noindent \textbf{Loss Design.} The established understanding is that the inputs injected into coarse layers play a dominant role in abstracting high-level features, encompassing global structure, viewpoint, large-scale attributes, and object shape. 
Utilizing this fact, the entire frequency encoding module is trained to ensure that the generated outputs maintain similarities to the characteristics of the guiding image by minimizing the associated loss function. Denoting the output vector of the CFM as $\mathbf{q}$, where $\mathbf{q} = \mathbf{q}^{\scriptscriptstyle 4}$, the encoding module aims to align the content-guiding vector 
closely match to $\mathbf{w}_{\scriptscriptstyle 1}$, while allowing perturbations through a randomly generated vector $\mathbf{w}_{\scriptscriptstyle 2}$:  
\begin{align}
    \mathcal{L} =  
    \frac{1}{N_b}\sum\limits^{N_b}_{j=1} 
    \lvert \lvert \mathbf{q}_{\scriptscriptstyle j} - \mathbf{w}_{\scriptscriptstyle 1, j} \rvert \rvert_{\scriptscriptstyle 2}^{\scriptscriptstyle 2}.
\end{align}

In the inference process, a real image $\mathbf{y}$ is used as input instead of generated images $\mathbf{x}_{\scriptscriptstyle f}$. The guiding vector $\mathbf{q}$ is injected into the coarse blocks of the generator, and the style vector $\mathbf{w}$ is inputted into the remaining blocks. This approach enables the resulting image to capture attribute features from the input image (\eg shape, age, and hairstyle in face image generation) while maintaining a similar-looking background or color distribution as the image generated by inputting $\mathbf{w}$ into all layers.

\begin{figure}[t]
\begin{center}
\begin{tabular}{ccc}
\includegraphics[width = 0.21\textwidth]{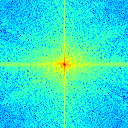} &
\includegraphics[width = 0.21\textwidth]{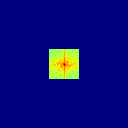} &
\includegraphics[width = 0.21\textwidth]{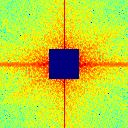}\vspace{-0.1cm}\\
{\footnotesize (a)} &
{\footnotesize (b)} &
{\footnotesize (c)} \vspace{0.1cm}\\
\includegraphics[width = 0.21\textwidth]{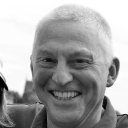}&
\includegraphics[width = 0.21\textwidth]{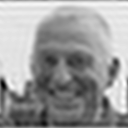}& 
\includegraphics[width = 0.21\textwidth]{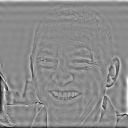}\\
{\footnotesize (d)} &
{\footnotesize (e)} &
{\footnotesize (f)}\\
\end{tabular}
\caption{Examples of different composition factors according to different frequency bands. (a) Magnitude of DFT. DC is in the middle. (b) Low-pass~($Y_{\scriptscriptstyle L}$) filtered DFT with $b_{\scriptscriptstyle L} = 30$ of (a). (c) High-pass~($Y_{\scriptscriptstyle H}$) filtered DFT with $b_{\scriptscriptstyle H} = 30$ of (a). (d) Inverse DFT of size $128 \times 128$ of (a). (e) Inverse DFT of (b). The overall layout of the original image is preserved, while the boundary regions are smoothed. (f) Inverse DFT of (c). This filter highlights the edges in the image while suppressing the homogeneous components.} \label{fig:frequency_analysis} 
\end{center}
\end{figure}

\noindent \textbf{Frequency Selection.}  
There is a large amount of evidence that the frequency component is a crucial factor in interpreting images~\citep{oliva2006hybrid, petras2019coarse, wang2020high, lee2022lossless, convFreq, imageConv2FreqConv, waveletGAN, yin2019fourier, schwarz2021frequency}. These findings have been established through frequency component analysis, as exemplified in Figure~\ref{fig:frequency_analysis}. Acknowledging the distinct influence of different frequency components on image composition, as observed in the figure, we posit that leveraging specific frequency components can offer advantages in controlling the desired content and the quality of generated images. Subsequently, we conduct an analysis to examine this hypothesis. This involves integrating frequency information into the generation process and studying the impact of different frequency bands. As a result, we anticipate potential enhancement of the generation process that better preserves desired content characteristics and improves the overall quality of the generated images. This utilization of frequency components opens up new possibilities for more effective and precise control over the generated image content and overall image generation process.

To scrutinize our hypothesis, we employ the two-dimensional discrete Fourier transform~(DFT) to extract frequency features, which is a widely used tool for frequency analysis. The DFT is a transform that converts a finite sequence of samples into a representation that describes frequency components.
Given an intensity~(grayscale) image $\widetilde{\mathbf{x}}\left(m,n\right) \in \mathbb{R}^{M \times N}$ converted from $\mathbf{x}$, followed by the definition of luminance in ~\citep{series2011studio}, the two-dimensional DFT $\:\mathbf{X}: \mathbb{R}^{M \times N} \rightarrow \mathbb{C}^{M \times N}$ is defined as follows~\citep{gonzalez2009digital}:
\begin{align}\label{eq:dft}
      \mathbf{X}\left(u, v\right) = \sum_{m=0}^{M-1} \sum_{n=0}^{N-1} \widetilde{x}\left(m,n\right) e^{-j2\pi (\frac{um}{M} + \frac{vn}{N})}, 
 \end{align}
where $u = 0, 1, \cdots, M-1$ and $v = 0, 1, \cdots, N-1$ denote equally spaced frequency variables, and $\left(m, n\right)$ denotes a spatial location of $\widetilde{\mathbf{x}}$. To reduce storage and bandwidth requirements, we resize the input intensity image to half its original size, resulting in $\widetilde{\mathbf{x}} \in \mathbb{R}^{128 \times 128}$, where $M$ and $N$ equal to $128$. This resizing operation has been found to have no significant impact on the model's performance.

Each computed DFT component in (\ref{eq:dft}) corresponds to a specific frequency, which is determined by its position $u$ and $v$. The magnitude of each component represents the strength or intensity of that frequency in $\widetilde{\mathbf{x}}$. Multiplying the magnitude of each DFT component by $(-1)^{m+n}$ shifts the direct current~(DC) component $X(0,0)$ to the center $(\lfloor M/2 \rfloor, \: \lfloor N/2 \rfloor)$ of the frequency spectrum. The shifted DFT is then denoted as $\mathbf{Y}(k,\ell) = \mathbf{X}(u-\lfloor M/2 \rfloor, v-\lfloor N/2 \rfloor)$, where $k$ and $\ell$ represent both positive and negative frequencies, \ie $k = -\lfloor M/2 \rfloor, \cdots, \lceil M/2 \rceil-1$ and $\ell = -\lfloor N/2 \rfloor, \cdots, \lceil N/2 \rceil-1$. In the experiments of this paper, we used this shifted version of the frequency spectrum.

To conduct a more rigorous analysis of the influence of frequency components across different bands, the frequency spectrum denoted as $\mathbf{Y}_{f}$ can be adjusted by selectively reducing specific frequency components, where the index $f \in \left\{ L, H\right\}$ denotes the low-pass~($L$) and high-pass~($H$) filtering operations, defined as follows:

\begin{align}\label{eq:lowpass}
        \mathbf{Y}_{L}\left(u,v\right)\!=\!\left\{\!
        \begin{array}{cl}
        \mathbf{Y}\left(u,v\right), \!\!&\!\!\textrm{if} \:\left| u-u_c\right| \leq b_{\scriptscriptstyle L} \land \left| v-v_c\right| \leq b_{\scriptscriptstyle L}\\
        0, & \textrm{otherwise}, 
        \end{array} \right. 
\end{align} \vspace{-0.3cm}
\begin{align}\label{eq:highpass}
        \mathbf{Y}_{H}\left(u,v\right)\!= \!\left\{\!
        \begin{array}{cl}
        \mathbf{Y}\left(u,v\right), \!\!&\!\!\textrm{if} \: \left| u-u_c\right| \geq b_{\scriptscriptstyle H} \land \left| v-v_c\right| \geq b_{\scriptscriptstyle H}\\
        0, & \textrm{otherwise},
        \end{array} \right.
\end{align}
where $u_c$ and $v_c$ are the positions of zero frequency; and the hyper-parameters $b_{\scriptscriptstyle L}$ and $b_{\scriptscriptstyle H}$ denote the cutoff values for highlighting some range of frequencies. 

Once frequency characteristics change by (\ref{eq:lowpass}) or (\ref{eq:highpass}), the frequency-selected image $\mathbf{x}_{\scriptscriptstyle f}\left(m, n\right) \vspace{0.1cm}$ is acquired by the inverse DFT, $\:\mathbf{X}^{-1}: \mathbb{C}^{M \times N} \rightarrow \mathbb{R}^{M \times N}$ to preserve both spatial information and refined frequency information:
\begin{align}\label{eq:idft}
     \mathbf{x}_{\scriptscriptstyle f}\left(m, n\right) = \frac{1}{MN} \sum_{u=0}^{M-1} \sum_{v=0}^{N-1} \mathbf{Y}_{f}\left(u,v\right) e^{j2\pi (\frac{mu}{M} + \frac{nv}{N})},     
 \end{align}
for $m = 0, 1, \cdots, M-1$ and $n = 0, 1, \cdots, N-1$. 

The impacts of frequency selection in image composition by (\ref{eq:lowpass}) and (\ref{eq:highpass}) are visually demonstrated in Figure~\ref{fig:frequency_analysis}. The figure comprehensively depicts how different frequency bands emphasize specific characteristics when composing images. This visualization provides compelling evidence for the significant impact of frequency components in image composition. In particular, the inverse DFT image in (b) preserves low-frequency components by utilizing a filter described by (\ref{eq:lowpass}), smoothing some details compared to the original images. On the other hand, the inverse DFT image in (c) highlights edges by applying a filter based on (\ref{eq:highpass}), suppressing homogeneous components in the original image. Based on these observations and theoretical reasoning~\citep{strang1996wavelets, gonzalez2009digital, lee2018camera}, we can prioritize low-frequency components that are directly related to the content of images, while suppressing high-frequency components that are mainly related to apperance. This approach allows us to focus on and transfer only the content of the image, providing a strong rationale for using frequency analysis in image generation. By emphasizing content and minimizing stylistic influence, frequency analysis and filtering improve content transfer and reduce stylistic interference. These advantages are further supported by the experimental results presented in Section~\ref{sec:results}.

\section{Results and Discussion}\label{sec:results}
This section presents experimental results based on the proposed method described in Section~\ref{sec:proposed_method}. An overview of the training dataset is provided in Section~\ref{sec:datasets}, followed by a discussion of the implementation details in Section~\ref{sec:implementation}. The effectiveness of the proposed framework is evaluated and demonstrated through the experimental results presented in Section~\ref{sec:experimental_results}. Finally, Section~\ref{sec:analysis} provides an in-depth analysis, exploring the effects of frequency analysis on image generation and performance comparison metrics.

\subsection{Datasets}\label{sec:datasets}
The effectiveness of the proposed framework was evaluated using four benchmark datasets commonly used in image generation tasks: FFHQ~\citep{stylegans}, AFHQ Cat and Dog~\citep{starganv2}, and LSUN Church and Bedroom~\citep{lsun}. Additionally, the CelebFaces Attributes \\
(CelebA)~\citep{celebA} dataset was utilized to train binary classification models and demonstrate the efficiency of the proposed approach. 

\noindent \textbf{FFHQ} comprises $70,000$ high-quality images at high-definition~(HD) resolution~($1024 \times 1024$), making it one of the largest datasets of its kind. It is widely used for its diverse collection of facial features, including variations in gender, age, and geographic origin. The dataset also includes a wide range of accessories such as glasses, necklaces, hats, earrings, and cosplay costumes. The FFHQ dataset also offers a variety of backgrounds, further enhancing the dataset's richness and making it a good choice for researchers aiming to train machine-learning models that can recognize and synthesize facial features in a realistic manner.

\noindent \textbf{AFHQ} is a comprehensive and high-quality collection of animal face images. It contains a diverse range of breeds, ages, poses, and accessories.
The images in the AFHQ dataset are of high resolution, with each image having dimensions of $512 \times 512$ pixels. 
The dataset consists of three domains: cat, dog, and wildlife. Each domain contains nearly $5,000$ images. However, for the study of this paper, the cat and dog domains were used. To ensure consistency and quality, the images in the AFHQ dataset were carefully processed, including vertical and horizontal alignment to center the eyes. This pre-processing helps to ensure uniformity in the images and facilitates accurate analysis and modeling.

\noindent \textbf{LSUN} is a large-scale collection of scene images consisting of $10$ categories, including dining rooms, bedrooms, churches, and classrooms. It provides a substantial amount of training data, with each category containing approximately $120,000$ to $3,000,000$ images. Additionally, there is a small validation set consisting of around $300$ images and a test set of $1,000$ images for each category. For evaluation purposes, we focused on the bedroom and outdoor church categories. To train the GANs, we combined the train, validation, and test sets of these categories. The outdoor church dataset contains $126,227$ images, which is a reasonable size for training GANs. However, the bedroom dataset is much larger, containing over $3$ million images. To create a more manageable dataset, we randomly selected a subset of $100,000$ images from the original set. By selecting a representative subset of images, we aimed to strike a balance between dataset size and computational feasibility, ensuring that the training process remained efficient and effective.

\noindent \textbf{CelebA} is a widely used dataset that consists of a comprehensive collection of $202,599$ face images belonging to $10,177$ different celebrities. Each image in the dataset is annotated with $40$ binary labels, providing information about various facial attributes such as hair color, gender, age, and more. This extensive annotation makes the CelebA dataset valuable for training and evaluating algorithms related to face recognition, facial attribute prediction, and face synthesis. 
In the experiments of this paper, the CelebA dataset was specifically used to train the face attribute prediction model. By leveraging the vast amount of labeled facial images in the dataset, we aimed to develop a robust and accurate method for evaluating content preservation.

\subsection{Implementation Details}\label{sec:implementation} 
The proposed framework was implemented using the PyTorch, version 1.13.1, CUDA Toolkit 11.7, and cuDNN 8.8.0 on a single NVIDIA GeForce RTX 3090 GPU with 24GB VRAM.

During the training of the generating module in Section~\ref{sec:generating_module}, we followed the training strategy outlined in StyleGAN2~\citep{stylegans2} without data augmentation. 
To ensure stable training, we applied the equalized learning rate technique~\citep{stylegans} to all trainable parameters. A leaky ReLU activation~\citep{ReLU} with a negative slope of $0.2$ was used, and a minibatch standard deviation layer was included at the end of the discriminator~\citep{stylegans}. For optimization, we employed the Adam optimizer~\citep{adam} with a learning rate of $0.0002$ and $\beta_1=0$, $\beta_2=0.99$. The minibatch size was set to $32$, and the exponential moving average (EMA) was used on the generator and mapping network parameters. The EMA parameters were updated once every 10-generator iteration. 

The frequency encoding module described in Section~\ref{sec:frequency_encoding_module} was trained using the pre-trained models $E_\mathbf{\psi}$ and $G_\mathbf{\phi}$ from the generating module with $\mathbf{\psi}$ and $\mathbf{\phi}$ being frozen. During the training of this module, we applied batch normalization with a momentum value of $0.9$. Each model was trained for a total of $50$ epochs, using a batch size of $32$. The Adam optimizer was utilized with parameters $\beta_1=0.9$, $\beta_2=0.999$, and a learning rate of $0.0001$. The depth of each encoder module progressively increased in the following order: $\left[128, 256, 512, 512\right]$.

\subsection{Experimental Results} \label{sec:experimental_results} 
In order to validate the effectiveness of the proposed framework, we conducted an assessment from two perspectives: 1) distribution fidelity, which measures the similarity between the distribution of generated images and real images, and 2) content fidelity, which quantifies how much the generated images maintain the contents of the guided images.
\subsubsection{Distribution Fidelity} \label{sec:similarity_results}
To evaluate the similarity of the generated image distribution with real images, the performance of the proposed framework was compared with that of the
that of the current state-of-the-art GAN model, StyleGAN2~\citep{stylegans2}, as well as with diffusion models including ADM~\citep{dhariwal2021diffusion} and Diffusion-GAN~\citep{wang2023diffusion}. Note that since StyleGAN2 does not provide evaluation results under the same configuration as the proposed framework, we re-trained StyleGAN2 using the configuration detailed in Section~\ref{sec:implementation} to ensure a fair comparison. For this purpose, we utilized the official PyTorch implementation of StyleGAN2-ADA~\citep{styleganADA}, developed by the same authors as StyleGAN2~\citep{stylegans2}, which has been confirmed to produce results consistent with the TensorFlow version. 

The performance of models was primarily evaluated using Precision~(P)
\citep{pr} on a set of $50,000$ generated images for each dataset. Precision measures the percentage of generated images that accurately match specific real images in the dataset, indicating how realistic the generated images are. Therefore, Precision is the most suitable metric for assessing the content similarity between generated and reference images, aligning with the objectives of this work. Nevertheless, Fréchet Inception Distance~(FID)~\citep{fid} scores were also provided, as it is a commonly used metric in image generation. FID focuses more on quantifying the diversity of the generated images compared to the desired distribution~\citep{pr}. 

\begin{table}[t]
    \centering
    \caption{Performance comparisons on FFHQ~\citep{stylegans}, AFHQ Cat and Dog~\citep{starganv2}, and LSUN Bedroom and Church~~\citep{lsun} in terms of Fréchet Inception Distance~(FID) and Precision~(P). Precision measures the percentage of generated images that accurately match specific real-world images in the dataset, offering insight into the realism of the generated images. Therefore, higher Precision values imply that a model can produce images preserving the content of reference images. \vspace{0.2cm}} \label{tab:evaluation}
    \renewcommand{\tabcolsep}{1mm}{ 
    {\renewcommand{\arraystretch}{1.2}
        \begin{tabular}{lcccccccc}
            \hline
             \multirow{2}{*}{Dataset} & \multicolumn{2}{c}{StyleGAN2} & \multicolumn{2}{c}{ADM} & \multicolumn{2}{c}{Diffusion-GAN} &  \multicolumn{2}{c}{Proposed} \\ \cline{2-3} \cline{4-5} \cline{6-7} \cline{8-9}  
             & \cellcolor{gray!10} P $\uparrow$ & FID $\downarrow$ & \cellcolor{gray!10} P $\uparrow$ & FID $\downarrow$ & \cellcolor{gray!10} P $\uparrow$ & FID $\downarrow$ & \cellcolor{gray!10} P $\uparrow$ & FID $\downarrow$ \\ \hline
             
             FFHQ & 
             \cellcolor{gray!10} 0.694 & 4.26 & \cellcolor{gray!10} 0.708 & 22.18 & \cellcolor{gray!10} 0.689 & 2.83 &  
             \cellcolor{gray!10} 0.812 & 15.2  \\  
             
             AFHQ Cat &  
             \cellcolor{gray!10} 0.710 & 5.29 & \cellcolor{gray!10} 0.788 & 12.65 & \cellcolor{gray!10} 0.612 & 2.4 &    
             \cellcolor{gray!10} 0.823 & 12.52   \\ 
             
             AFHQ Dog & 
             \cellcolor{gray!10} 0.653 & 29.92 & \cellcolor{gray!10} 0.822 & 20.42 & \cellcolor{gray!10} 0.600 & 4.83 &  
             \cellcolor{gray!10} 0.814 & 30.84  \\
             
             LSUN Bedroom & 
             \cellcolor{gray!10} 0.547 & 3.65 & \cellcolor{gray!10} 0.660 & 1.9 & \cellcolor{gray!10} 0.606 & 3.65 & 
             \cellcolor{gray!10} 0.690 & 14.34   \\  
             
             LSUN Church & 
             \cellcolor{gray!10} 0.573 & 3.35 & \cellcolor{gray!10} 0.496 & 21.09 & \cellcolor{gray!10} 0.573 & 3.17 &  
             \cellcolor{gray!10} 0.774 & 10.72  \\ 
             \hline
        \end{tabular}
    }} 
\end{table}

This evaluation results are provided in Table~\ref{tab:evaluation}. It is important to note that in the experiments conducted in this section, a lowpass filter with a cutoff frequency of $b_{\scriptscriptstyle L} = 10$ was applied to all test datasets. The filter type and cutoff value were selected based on the experiments described in Section~\ref{sec:extensive_frequency}, where they provided the best precision suitable for the task. According to the results in Table~\ref{tab:evaluation}, the precision of the proposed framework showed significant improvements compared to StyleGAN: $17\%$ on FFHQ, $16\%$ on AFHQ Cat, $25\%$ on AFHQ Dog, $20\%$ on LSUN Bedroom, and $35\%$ on LSUN Church. Additionally, performance improved by approximately  $15\%$ on FFHQ, $4\%$ on AFHQ Cat, $5\%$ on LSUN Bedroom, and $56\%$ on LSUN Church compared to ADM. Compared to Diffusion-GAN, performance improved by approximately  $18\%$ on FFHQ, $34\%$ on AFHQ Cat, $36\%$ on AFHQ Dog, $14\%$ on LSUN Bedroom, and $36\%$ on LSUN Church. In contrast, the FID scores of the proposed framework were higher than those of the comparison method. This discrepancy arose because the proposed method emphasizes preserving the attributes of reference images, resulting in generated images that closely resemble specific images in the real dataset. As a consequence, the proposed framework might not fully capture the overall statistical properties of the real dataset. This highlights the trade-off between similarity and diversity in the training of GANs. Further analysis of the tradeoff relationship between Precision and FID in Section~\ref{sec:extensive_tradeoff} suggests that as Precision increases, FID inevitably increases. Therefore, higher Precision values confirm that the proposed framework can produce images preserving content attributes of reference images, even as FID increases, aligning with our intended goal.

\begin{figure*}[t]
  \begin{minipage}[c]{0.48\textwidth}
      \centering
      \begin{tabular}{@{}c@{~}c@{~}|@{~}c@{~}c@{~}c@{~}c@{~}c@{}}
        & \multicolumn{5}{c}{\hspace{2.5cm} \footnotesize $\mathbf{w}_{\scriptscriptstyle 2}$ } \\
        
        \multicolumn{2}{c}{}  &
        \includegraphics[width = 0.14\textwidth]{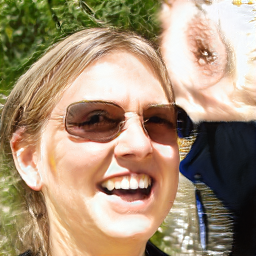} &
        \includegraphics[width = 0.14\textwidth]{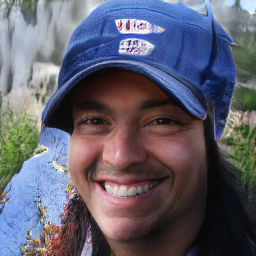} &
        \includegraphics[width = 0.14\textwidth]{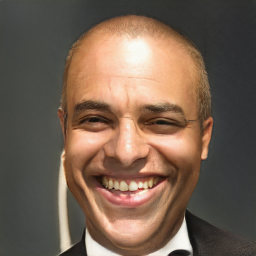} &
        \includegraphics[width = 0.14\textwidth]{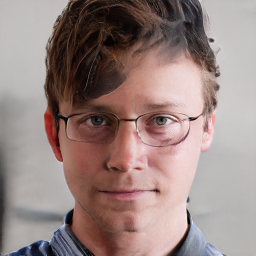} &
        \includegraphics[width = 0.14\textwidth]{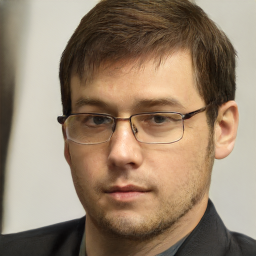} \\
        \cline{3-7} \noalign{\vskip 0.15cm}
        
        &
        \includegraphics[width = 0.14\textwidth]{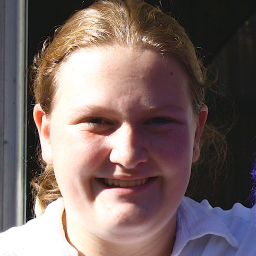} &
        \includegraphics[width = 0.14\textwidth]{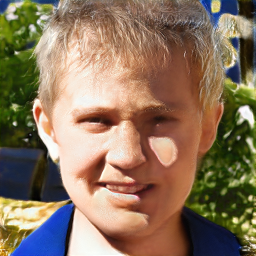} &
        \includegraphics[width = 0.14\textwidth]{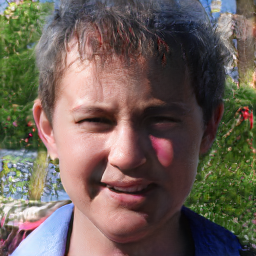} &
        \includegraphics[width = 0.14\textwidth]{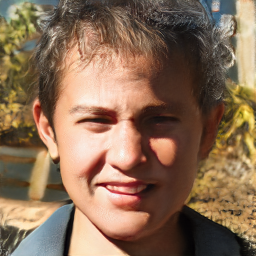} &
        \includegraphics[width = 0.14\textwidth]{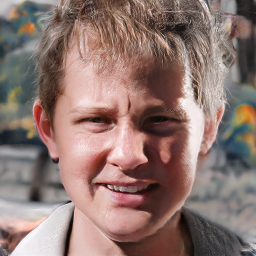} &
        \includegraphics[width = 0.14\textwidth]{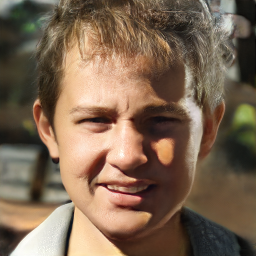} \\
    
        \multirow{5}{*}{\rotatebox{90}{\footnotesize Real}} &
        \includegraphics[width = 0.14\textwidth]{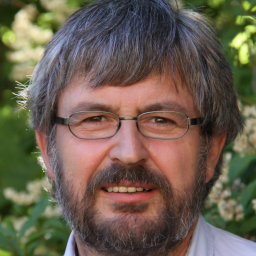} &
        \includegraphics[width = 0.14\textwidth]{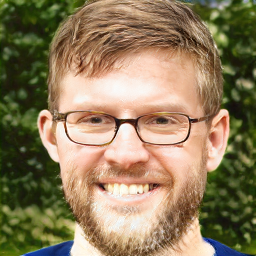} &
        \includegraphics[width = 0.14\textwidth]{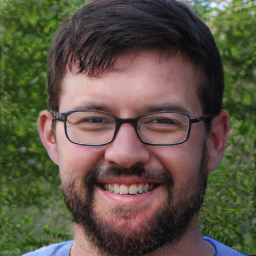} &
        \includegraphics[width = 0.14\textwidth]{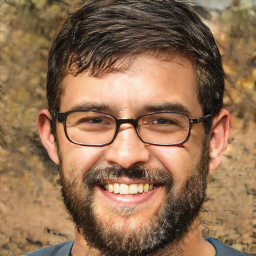} &
        \includegraphics[width = 0.14\textwidth]{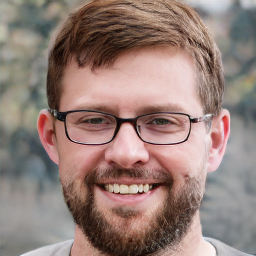} &
        \includegraphics[width = 0.14\textwidth]{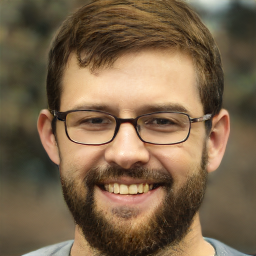} \\
    
        &
        \includegraphics[width = 0.14\textwidth]{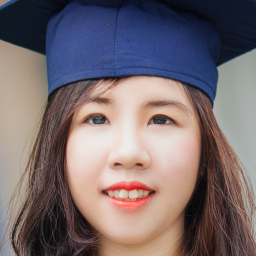} &
        \includegraphics[width = 0.14\textwidth]{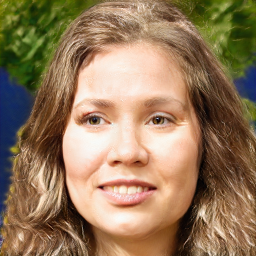} &
        \includegraphics[width = 0.14\textwidth]{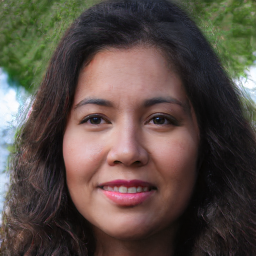} &
        \includegraphics[width = 0.14\textwidth]{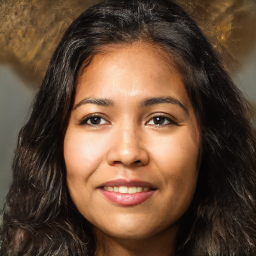} &
        \includegraphics[width = 0.14\textwidth]{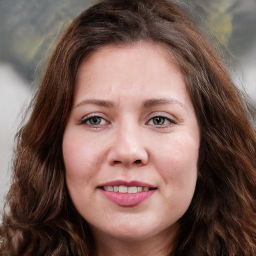} &
        \includegraphics[width = 0.14\textwidth]{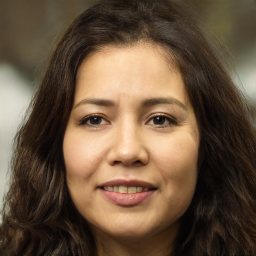} \\
    
        &
        \includegraphics[width = 0.14\textwidth]{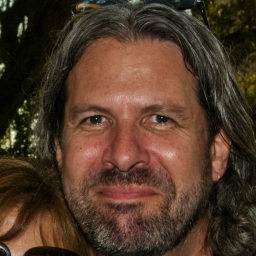} &
        \includegraphics[width = 0.14\textwidth]{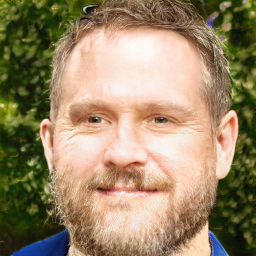} &
        \includegraphics[width = 0.14\textwidth]{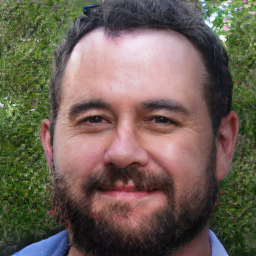} &
        \includegraphics[width = 0.14\textwidth]{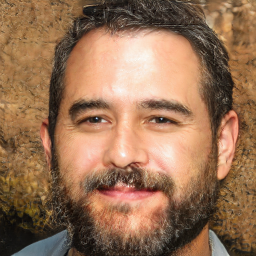} &
        \includegraphics[width = 0.14\textwidth]{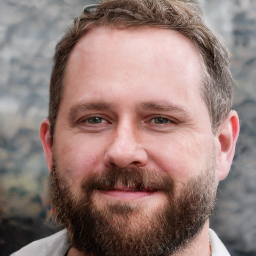} &
        \includegraphics[width = 0.14\textwidth]{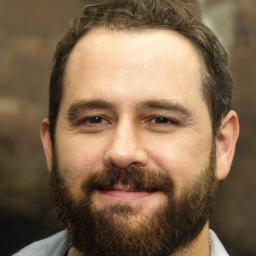} \\
      \end{tabular} 
      \\ \vspace{0.05cm}{\footnotesize (a)}
    \end{minipage} 
    {\hskip 0.1cm}
    \begin{minipage}[c]{0.45\textwidth}
      \centering
      \vspace{0.4cm}
      {\footnotesize
      \begin{tabular}{c@{}c@{~}|@{~}c@{}@{~}c@{}@{~}c@{}@{~}c@{}@{~}c@{}}
            \multicolumn{2}{c}{} &
            \framebox(1.0cm, 1.0cm){\makecell{
             \tiny{A}, \vspace{-0.1cm}\\ \tiny{DC}, \vspace{-0.1cm}\\ \tiny{BL},}} &
            
            \framebox(1.0cm, 1.0cm){\makecell{
            \tiny{Y},  \vspace{-0.1cm}\\ \tiny{WH},  \vspace{-0.1cm}\\ \tiny{BN}, }} &
            
            \framebox(1.0cm, 1.0cm){\makecell{
            \tiny{B},  \vspace{-0.1cm}\\ \tiny{BL},  \vspace{-0.1cm}\\ \tiny{NE}, }} &
            
            \framebox(1.0cm, 1.0cm){\makecell{
             \tiny{A}, \vspace{-0.1cm}\\ \tiny{Y},  \vspace{-0.1cm}\\ \tiny{PS},}} &
            
            \framebox(1.0cm, 1.0cm){\makecell{
            \tiny{Y}, \vspace{-0.1cm}\\ \tiny{SH},  \vspace{-0.1cm}\\ \tiny{BrH},}}
            \\ \cline{3-7} \noalign{\vskip 0.16cm}
            &
            \framebox(1.0cm, 1.0cm){\makecell{
            \tiny{NE},  \vspace{-0.1cm}\\ \tiny{NB},  \vspace{-0.1cm}\\ \tiny{S},}} &
            
            \framebox(1.0cm, 1.0cm){\makecell{
            \tiny{NE},  \vspace{-0.1cm}\\ \tiny{NB}, \vspace{-0.1cm}\\ \tiny{BUE},}} &
            
            \framebox(1.0cm, 1.0cm){\makecell{
            \tiny{NE},  \vspace{-0.1cm}\\ \tiny{NB}, \vspace{-0.1cm}\\ \tiny{BUE},}} &
            
            \framebox(1.0cm, 1.0cm){\makecell{
            \tiny{NE},  \vspace{-0.1cm}\\ \tiny{NB}, \vspace{-0.1cm}\\ \tiny{BUE},}} &
            
            \framebox(1.0cm, 1.0cm){\makecell{
            \tiny{NE},  \vspace{-0.1cm}\\ \tiny{NB}, \vspace{-0.1cm}\\ \tiny{BUE}, }} &
            
            \framebox(1.0cm, 1.0cm){\makecell{
            \tiny{BN},  \vspace{-0.1cm}\\ \tiny{NB}, \vspace{-0.1cm}\\ \tiny{BUE},}} \\
   
            &
            \framebox(1.0cm, 1.0cm){\makecell{
            \tiny{E},  \vspace{-0.1cm}\\ \tiny{BUE}, \vspace{-0.1cm}\\ \tiny{Sb},}} &

            \framebox(1.0cm, 1.0cm){\makecell{
            \tiny{E},  \vspace{-0.1cm}\\ \tiny{BUE}, \vspace{-0.1cm}\\ \tiny{S},}} &

            \framebox(1.0cm, 1.0cm){\makecell{
            \tiny{E},  \vspace{-0.1cm}\\ \tiny{BUE}, \vspace{-0.1cm}\\ \tiny{M},}} &

            \framebox(1.0cm, 1.0cm){\makecell{
            \tiny{E},  \vspace{-0.1cm}\\ \tiny{BUE}, \vspace{-0.1cm}\\ \tiny{S},}} &

            \framebox(1.0cm, 1.0cm){\makecell{
            \tiny{E},  \vspace{-0.1cm}\\ \tiny{BUE}, \vspace{-0.1cm}\\ \tiny{M},}} &

            \framebox(1.0cm, 1.0cm){\makecell{
            \tiny{E},  \vspace{-0.1cm}\\ \tiny{BUE}, \vspace{-0.1cm}\\ \tiny{M},}} \\
            
            &
            \framebox(1.0cm, 1.0cm){\makecell{
            \tiny{HC}, \vspace{-0.1cm}\\ \tiny{BrH}, \vspace{-0.1cm}\\ \tiny{HM},}} &

            \framebox(1.0cm, 1.0cm){\makecell{
            \tiny{HC}, \vspace{-0.1cm}\\ \tiny{BrH},  \vspace{-0.1cm}\\ \tiny{WL},}} &

            \framebox(1.0cm, 1.0cm){\makecell{
            \tiny{HC}, \vspace{-0.1cm}\\ \tiny{BH},  \vspace{-0.1cm}\\ \tiny{S},}} &

            \framebox(1.0cm, 1.0cm){\makecell{
            \tiny{HC}, \vspace{-0.1cm}\\ \tiny{BH},  \vspace{-0.1cm}\\ \tiny{S},}} &

            \framebox(1.0cm, 1.0cm){\makecell{
            \tiny{HC}, \vspace{-0.1cm}\\ \tiny{WaH},  \vspace{-0.1cm}\\ \tiny{WL},}} &

            \framebox(1.0cm, 1.0cm){\makecell{
            \tiny{HC}, \vspace{-0.1cm}\\ \tiny{WaH},  \vspace{-0.1cm}\\ \tiny{WL},}} \\
            
            &
            \framebox(1.0cm, 1.0cm){\makecell{
            \tiny{M}, \vspace{-0.1cm}\\ \tiny{BN}, \vspace{-0.1cm}\\ \tiny{Sb},}} &

            \framebox(1.0cm, 1.0cm){\makecell{      
            \tiny{M}, \vspace{-0.1cm}\\ \tiny{BN}, \vspace{-0.1cm}\\ \tiny{S},}} &

            \framebox(1.0cm, 1.0cm){\makecell{
            \tiny{M}, \vspace{-0.1cm}\\ \tiny{BN}, \vspace{-0.1cm}\\ \tiny{BH},}} &

            \framebox(1.0cm, 1.0cm){\makecell{
            \tiny{M}, \vspace{-0.1cm}\\ \tiny{BN}, \vspace{-0.1cm}\\ \tiny{BH},}} &

            \framebox(1.0cm, 1.0cm){\makecell{
            \tiny{M}, \vspace{-0.1cm}\\ \tiny{BN}, \vspace{-0.1cm}\\ \tiny{BrH},}} &

            \framebox(1.0cm, 1.0cm){\makecell{      
            \tiny{M}, \vspace{-0.1cm}\\ \tiny{BN}, \vspace{-0.1cm}\\ \tiny{BE},}} \\
        \end{tabular} }
        \\ {\footnotesize (b)}
    \end{minipage}
  \caption{Example of (a) generated images by the proposed framework trained on FFHQ~\citep{stylegans} and (b) their corresponding attributes determined by the classifiers described in Section~\ref{sec:content_results}. In (a), the images in the first column are real input images, and the images in the first row are the images generated from a projected vector $\mathbf{w}_{\scriptscriptstyle 2}$. These examples demonstrate the preservation of real image content in the row-wise direction and control over color distribution and background in the column-wise direction for generated images. The table in (b) displays examples of the classified attributes. Each cell in the table contains the abbreviations of attributes of the images corresponding to the positions in (a).} \label{fig:ffhq_results}
\end{figure*}

\begin{figure*}[t]
    \begin{minipage}[c]{0.49\textwidth}
      \centering
      \begin{tabular}{@{}c@{~}c@{~}|@{~}c@{~}c@{~}c@{~}c@{~}c@{}}
        & \multicolumn{5}{c}{\hspace{2.5cm} \footnotesize $\mathbf{w}_{\scriptscriptstyle 2}$ } \\
          
        \multicolumn{2}{c}{}  & 
        \includegraphics[width = 0.14\textwidth]{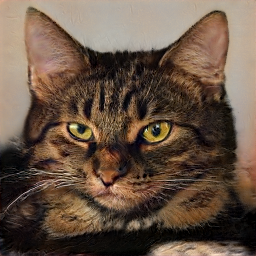} &
        \includegraphics[width = 0.14\textwidth]{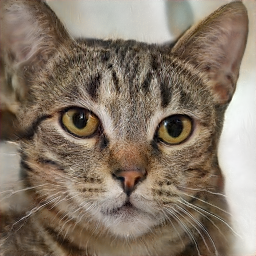} &
        \includegraphics[width = 0.14\textwidth]{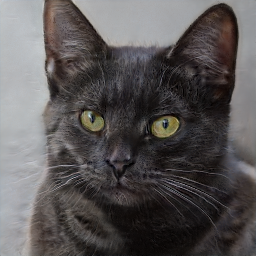} &
        \includegraphics[width = 0.14\textwidth]{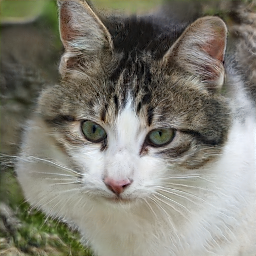} &
        \includegraphics[width = 0.14\textwidth]{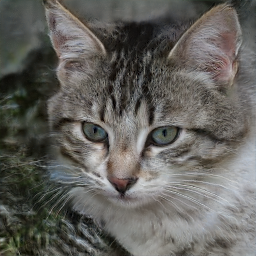} \\
        \cline{3-7} \noalign{\vskip 0.1cm}

        \multirow{3}{*}{\rotatebox{90}{\footnotesize Real}} &
        \includegraphics[width = 0.14\textwidth]{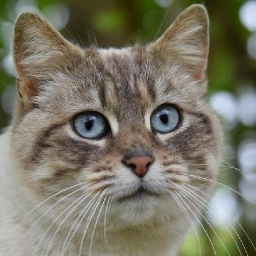} &
        \includegraphics[width = 0.14\textwidth]{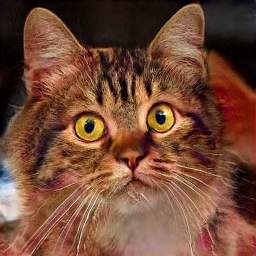} &
        \includegraphics[width = 0.14\textwidth]{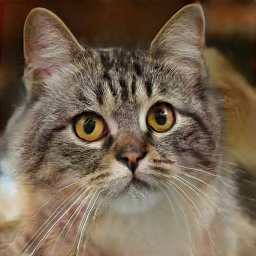} &
        \includegraphics[width = 0.14\textwidth]{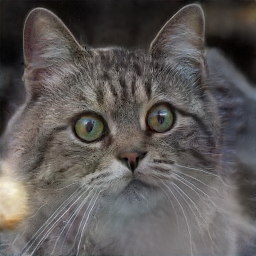} &
        \includegraphics[width = 0.14\textwidth]{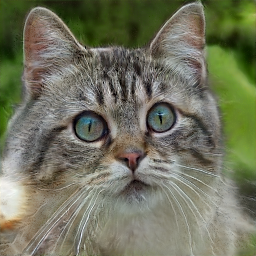} &
        \includegraphics[width = 0.14\textwidth]{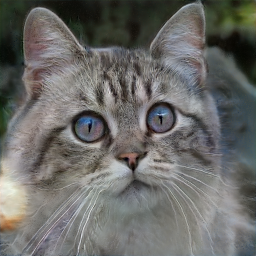} \\

        &
        \includegraphics[width = 0.14\textwidth]{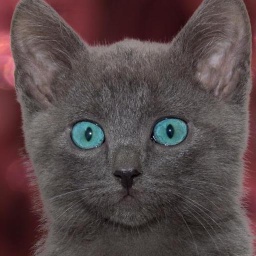} &
        \includegraphics[width = 0.14\textwidth]{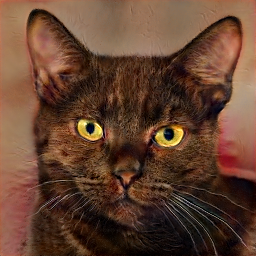} &
        \includegraphics[width = 0.14\textwidth]{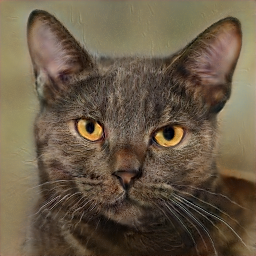} &
        \includegraphics[width = 0.14\textwidth]{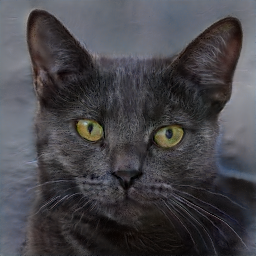} &
        \includegraphics[width = 0.14\textwidth]{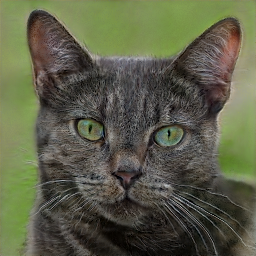} &
        \includegraphics[width = 0.14\textwidth]{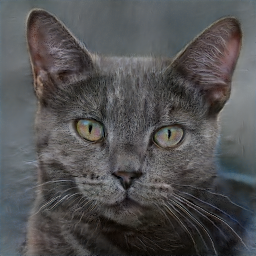} \\

        &
        \includegraphics[width = 0.14\textwidth]{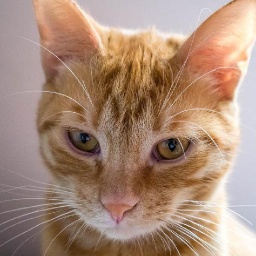} &
        \includegraphics[width = 0.14\textwidth]{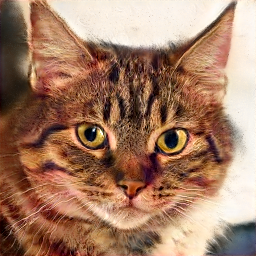} &
        \includegraphics[width = 0.14\textwidth]{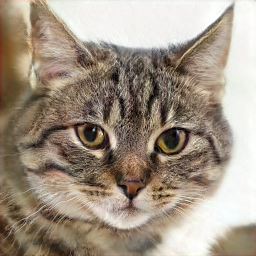} &
        \includegraphics[width = 0.14\textwidth]{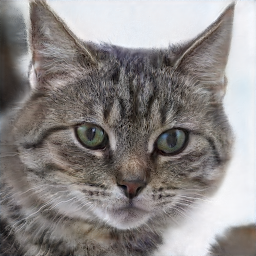} &
        \includegraphics[width = 0.14\textwidth]{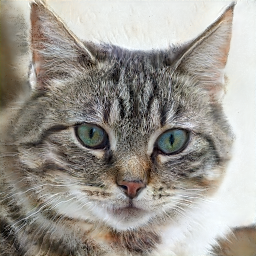} &
        \includegraphics[width = 0.14\textwidth]{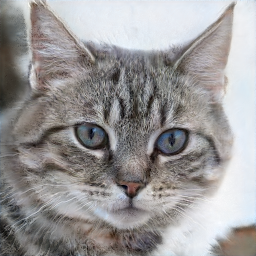} \\

        \multicolumn{7}{c}{\footnotesize Cat}
      \end{tabular}
    \end{minipage}     
    \begin{minipage}[c]{0.49\textwidth}
      \centering
      \begin{tabular}{@{}c@{~}c@{~}|@{~}c@{~}c@{~}c@{~}c@{~}c@{}}
        & \multicolumn{5}{c}{\hspace{2.5cm} \footnotesize $\mathbf{w}_{\scriptscriptstyle 2}$ } \\
    
        \multicolumn{2}{c}{}  & 
        \includegraphics[width = 0.14\textwidth]{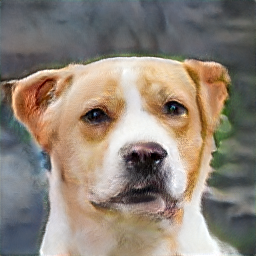} &
        \includegraphics[width = 0.14\textwidth]{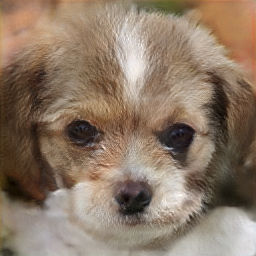} &
        \includegraphics[width = 0.14\textwidth]{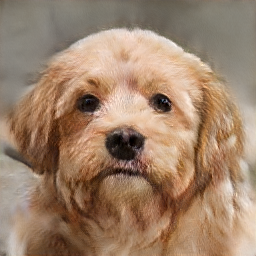} &
        \includegraphics[width = 0.14\textwidth]{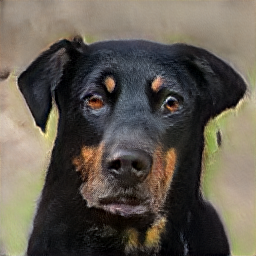} &
        \includegraphics[width = 0.14\textwidth]{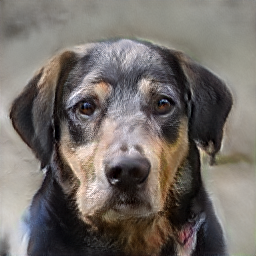} \\
        \cline{3-7} \noalign{\vskip 0.1cm}
        
        \multirow{3}{*}{\rotatebox{90}{\footnotesize Real}} &
        \includegraphics[width = 0.14\textwidth]{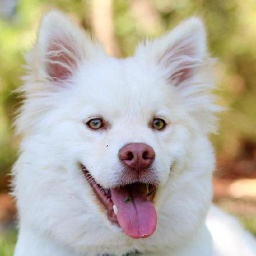} &
        \includegraphics[width = 0.14\textwidth]{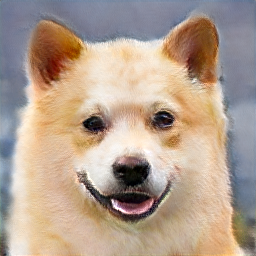} &
        \includegraphics[width = 0.14\textwidth]{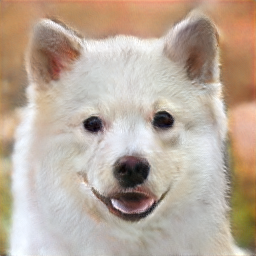} &
        \includegraphics[width = 0.14\textwidth]{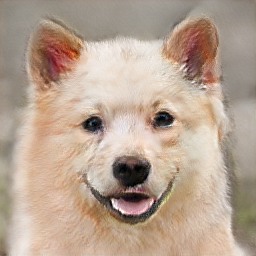} &
        \includegraphics[width = 0.14\textwidth]{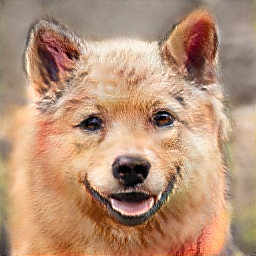} &
        \includegraphics[width = 0.14\textwidth]{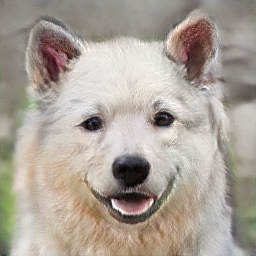} \\

        &
        \includegraphics[width = 0.14\textwidth]{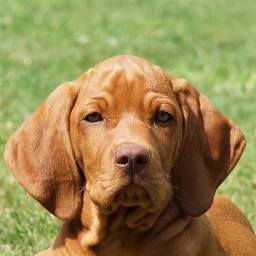} &
        \includegraphics[width = 0.14\textwidth]{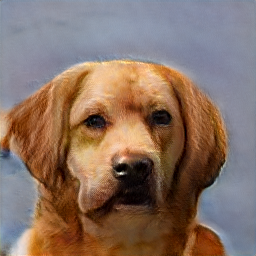} &
        \includegraphics[width = 0.14\textwidth]{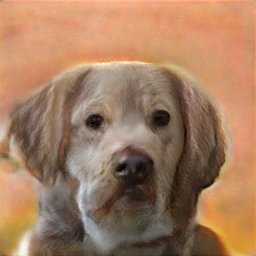} &
        \includegraphics[width = 0.14\textwidth]{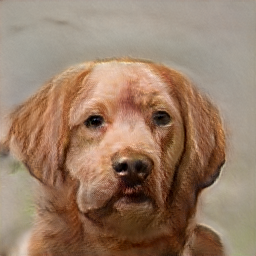} &
        \includegraphics[width = 0.14\textwidth]{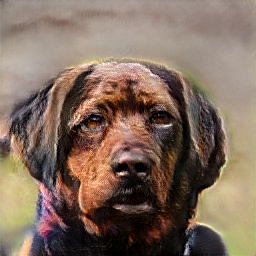} &
        \includegraphics[width = 0.14\textwidth]{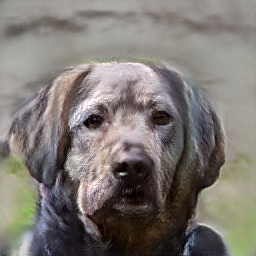} \\

        &
        \includegraphics[width = 0.14\textwidth]{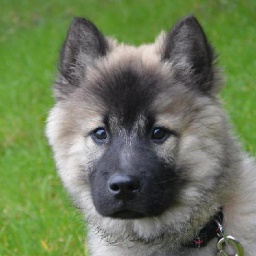} &
        \includegraphics[width = 0.14\textwidth]{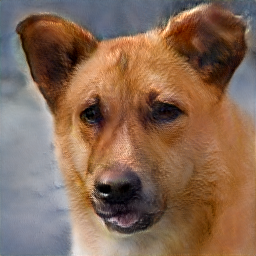} &
        \includegraphics[width = 0.14\textwidth]{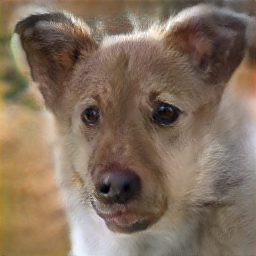} &
        \includegraphics[width = 0.14\textwidth]{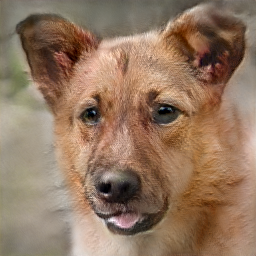} &
        \includegraphics[width = 0.14\textwidth]{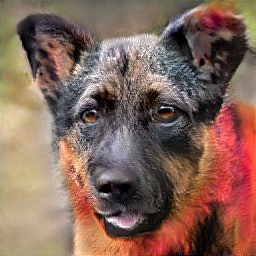} &
        \includegraphics[width = 0.14\textwidth]{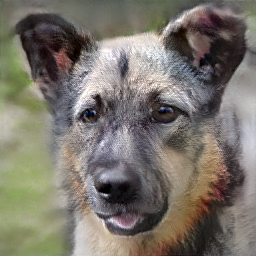} \\
        
        \multicolumn{7}{c}{\footnotesize Dog} 
      \end{tabular}
  \end{minipage}
  \label{fid:dog}
  \caption{Example of generated images by the proposed framework trained on AFHQ~\citep{starganv2}. The images in the first column are real input images, and the images in the first row are the images generated from a projected vector $\mathbf{w}_{\scriptscriptstyle 2}$ in Fig.~\ref{fig:inference}.} \label{fig:afhq_results}
\end{figure*}

\begin{figure*}[!h]
    \begin{minipage}[c]{0.49\textwidth}
      \centering
      \begin{tabular}{@{}c@{~}c@{~}|@{~}c@{~}c@{~}c@{~}c@{~}c@{}}
        & \multicolumn{5}{c}{\hspace{2.5cm} \footnotesize $\mathbf{w}_{\scriptscriptstyle 2}$ } \\

        \multicolumn{2}{c}{} & 
        \includegraphics[width = 0.14\textwidth]{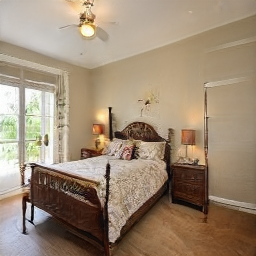} &
        \includegraphics[width = 0.14\textwidth]{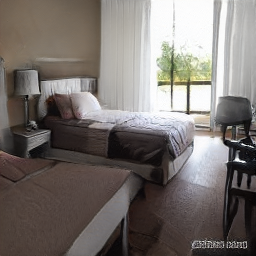} &
        \includegraphics[width = 0.14\textwidth]{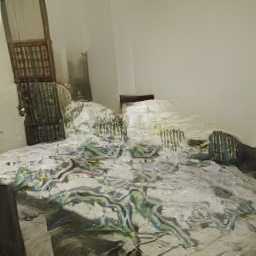} &
        \includegraphics[width = 0.14\textwidth]{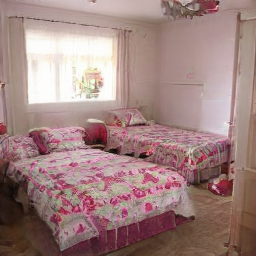} &
        \includegraphics[width = 0.14\textwidth]{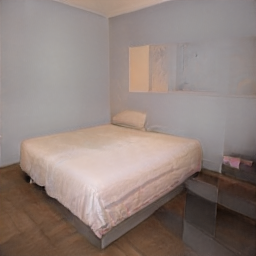} \\
        \cline{3-7} \noalign{\vskip 0.1cm}

        \multirow{3}{*}{\rotatebox{90}{\footnotesize Real}}&
        \includegraphics[width = 0.14\textwidth]{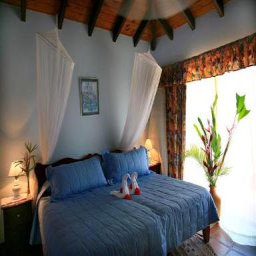} &
        \includegraphics[width = 0.14\textwidth]{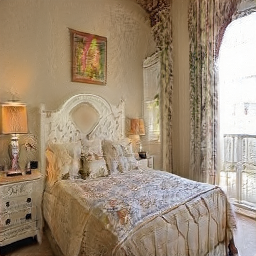} &
        \includegraphics[width = 0.14\textwidth]{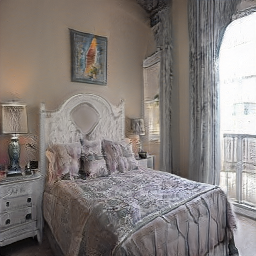} &
        \includegraphics[width = 0.14\textwidth]{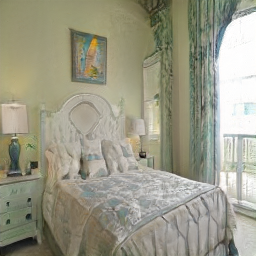} &
        \includegraphics[width = 0.14\textwidth]{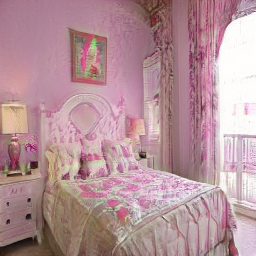} &
        \includegraphics[width = 0.14\textwidth]{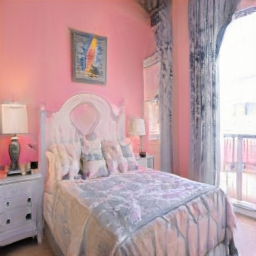} \\

        &
        \includegraphics[width = 0.14\textwidth]{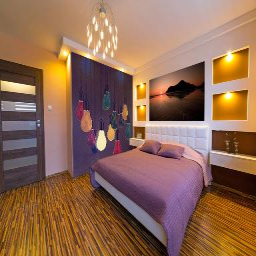} &
        \includegraphics[width = 0.14\textwidth]{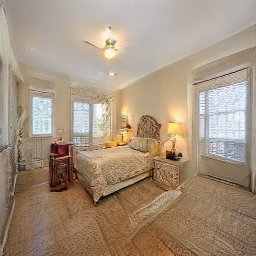} &
        \includegraphics[width = 0.14\textwidth]{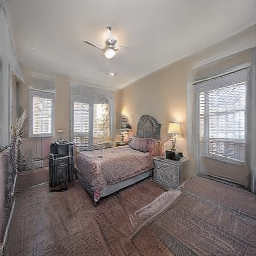} &
        \includegraphics[width = 0.14\textwidth]{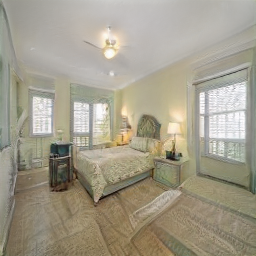} &
        \includegraphics[width = 0.14\textwidth]{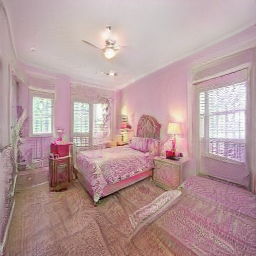} &
        \includegraphics[width = 0.14\textwidth]{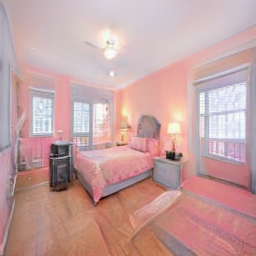} \\

        &
        \includegraphics[width = 0.14\textwidth]{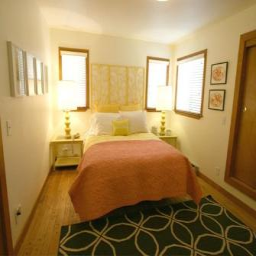} &
        \includegraphics[width = 0.14\textwidth]{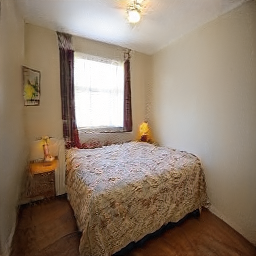} &
        \includegraphics[width = 0.14\textwidth]{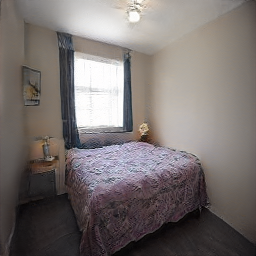} &
        \includegraphics[width = 0.14\textwidth]{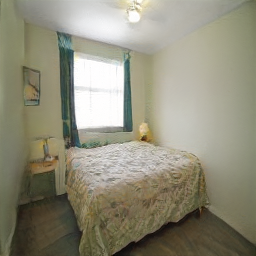} &
        \includegraphics[width = 0.14\textwidth]{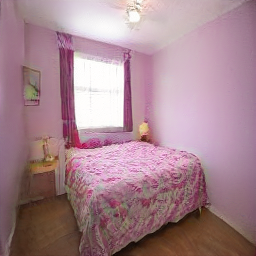} &
        \includegraphics[width = 0.14\textwidth]{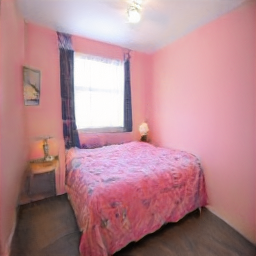} \\

        \multicolumn{7}{c}{\footnotesize Bedroom}
    \end{tabular}
  \end{minipage}
  \begin{minipage}[c]{0.49\textwidth}
      \centering
      \begin{tabular}{@{}c@{~}c@{~}|@{~}c@{~}c@{~}c@{~}c@{~}c@{}}
        & \multicolumn{5}{c}{\hspace{2.5cm} \footnotesize $\mathbf{w}_{\scriptscriptstyle 2}$ } \\
 
        \multicolumn{2}{c}{}  & 
        \includegraphics[width = 0.14\textwidth]{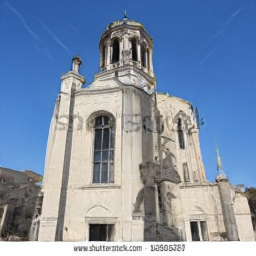} &
        \includegraphics[width = 0.14\textwidth]{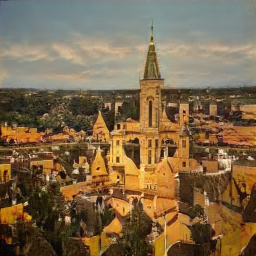} &
        \includegraphics[width = 0.14\textwidth]{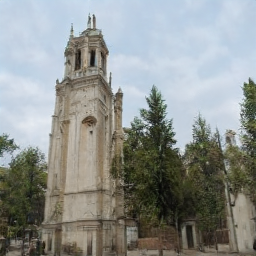} &
        \includegraphics[width = 0.14\textwidth]{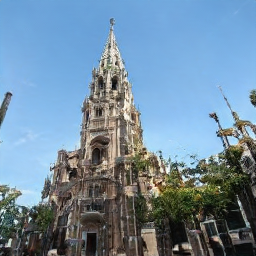} &
        \includegraphics[width = 0.14\textwidth]{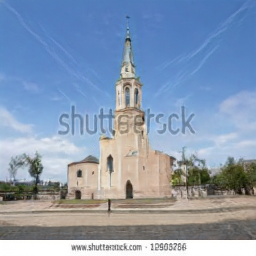} \\
        \cline{3-7} \noalign{\vskip 0.1cm}

        \multirow{3}{*}{\rotatebox{90}{\footnotesize Real}}& 
        \includegraphics[width = 0.14\textwidth]{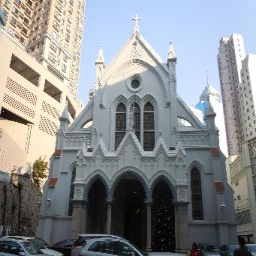} &
        \includegraphics[width = 0.14\textwidth]{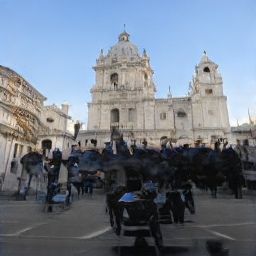} &
        \includegraphics[width = 0.14\textwidth]{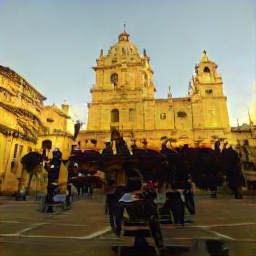} &
        \includegraphics[width = 0.14\textwidth]{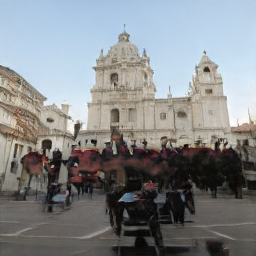} &
        \includegraphics[width = 0.14\textwidth]{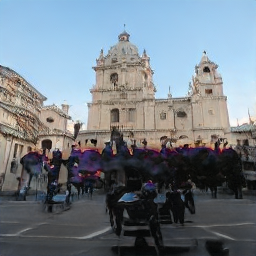} &
        \includegraphics[width = 0.14\textwidth]{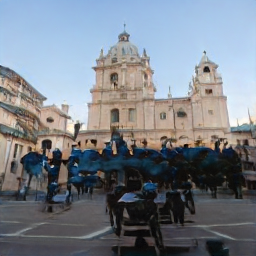} \\

        &
        \includegraphics[width = 0.14\textwidth]{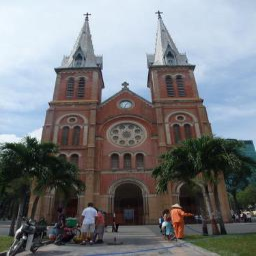} &
        \includegraphics[width = 0.14\textwidth]{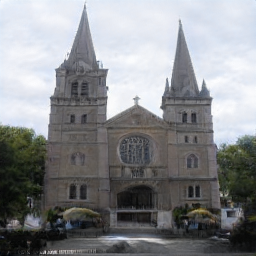} &
        \includegraphics[width = 0.14\textwidth]{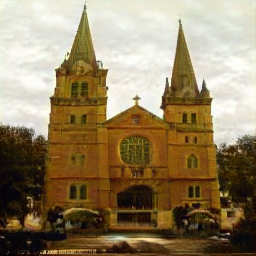} &
        \includegraphics[width = 0.14\textwidth]{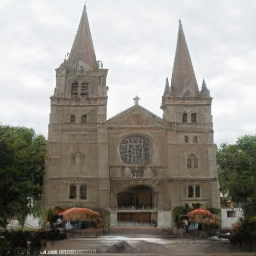} &
        \includegraphics[width = 0.14\textwidth]{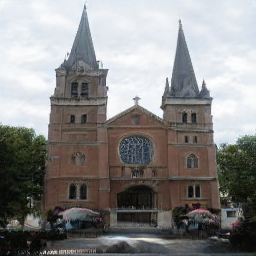} &
        \includegraphics[width = 0.14\textwidth]{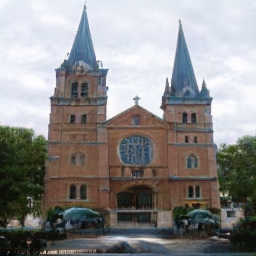} \\

        &
        \includegraphics[width = 0.14\textwidth]{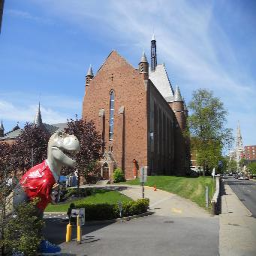} &
        \includegraphics[width = 0.14\textwidth]{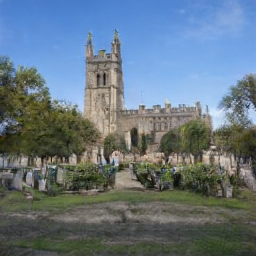} &
        \includegraphics[width = 0.14\textwidth]{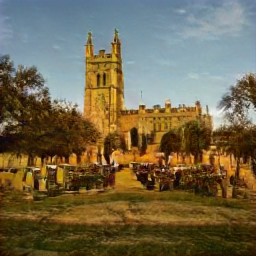} &
        \includegraphics[width = 0.14\textwidth]{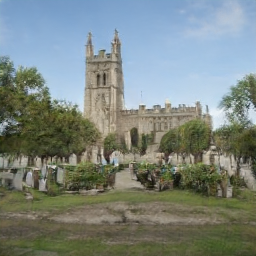} &
        \includegraphics[width = 0.14\textwidth]{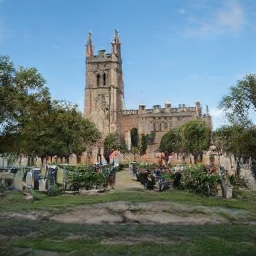} &
        \includegraphics[width = 0.14\textwidth]{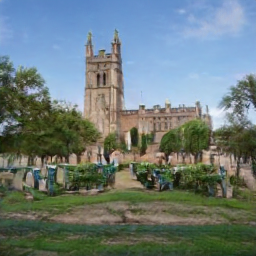} \\
        \multicolumn{7}{c}{\footnotesize Church}
     \end{tabular}
   \end{minipage}
  \caption{Example of generated images by the proposed framework trained on LSUN~\citep{lsun}. The images in the first column are real input images, and the images in the first row are the images generated from a projected vector $\mathbf{w}_{\scriptscriptstyle 2}$.}  \label{fig:lsun_results}
\end{figure*}

We also visually demonstrate the effectiveness of the proposed framework in Figs.~\ref{fig:ffhq_results}, \ref{fig:afhq_results}, and \ref{fig:lsun_results}. In the figures, the real input images are displayed in the first column, and the images generated from a projected vector $\mathbf{w}$ in Figure~\ref{fig:inference} are shown in the first row. The examples of the generated images obviously demonstrate that they preserved the content of the corresponding input real images within the same row. In particular, the generated face images Figure~\ref{fig:ffhq_results} exhibited similar attributes such as age, face shape, pose, direction, hairstyle, and accessories, while other attributes such as skin color, hair color, and ethnicity matched those in the first row. This observation indicates that the color distribution and backgrounds of the generated images were influenced by the injected vector $\mathbf{w}$ in the fine layers of $G_{\scriptscriptstyle \phi}$, as expected. 

As validated by the experiments, the proposed frequency encoding module effectively produces output images that resemble the content of the reference images, thereby providing users with control over the content of the generated images.

\subsubsection{Content Fidelity} \label{sec:content_results}
As specific evaluation metrics to quantify the transfer of content from real images to generated images are not available, we assessed the effectiveness of the proposed framework by examining its ability to preserve attributes of guiding images. However, since attribute labels for other benchmark datasets were unavailable, we employed the CelebA dataset to train attribute classifiers and applied these classifiers to the images of FFHQ. The CelebA dataset includes 40 facial attribute annotations, covering aspects such as gender, hairstyle,  and accessories, as detailed in Section~\ref{sec:datasets}. 

To conduct the experiment, we trained multiple binary classifiers for all attributes. These classifiers were then applied to both the guiding images and the generated images to determine the percentage of attributes that matched between them. 
The binary classifiers were implemented using EfficientNet-B4~\citep{efficientnet} and trained with the Adam optimizer. We set the optimizer's parameters as $\beta_1=0.9$, $\beta_2=0.999$, and a weight decay of 0.01. The training process lasted for $200$ epochs, using a batch size of $512$. To ensure a smooth learning rate transition, we employed a warm-up strategy, gradually increasing the learning rate from $0$ to $0.00001$~\citep{trainImageNetfast} until the first epoch. The training images were obtained by cropping to $90\%$ of their original size. Data augmentation techniques, such as horizontal flipping, random brightness and contrast adjustments, and cutout~\citep{cutout} with four holes~(each hole having a maximum edge length of $32$) were also applied.

The attribute classification results are exemplified for the output images of Figure~\ref{fig:ffhq_results}(a) in Figure~\ref{fig:ffhq_results}(b). Each cell of Figure~\ref{fig:ffhq_results}(b) contains abbreviations representing attributes corresponding to the images' positions in Figure~\ref{fig:ffhq_results}(a). The attributes include Attractive~(A), Bald~(B), Bushy Eyebrows~(BE), Black Hair~(BH), Big Lips~(BL), Big Nose~(BN), Brown Hair~(BrH), Bags Under Eyes~(BUE), Chubby~(C), Double Chin~(DC), Eyeglasses~(E), High Cheekbones~(HC), Heavy Makeup~(HM), Male~(M), Mouth Slightly Open~(MSO), No Beard~(NB), Narrow Eyes~(NE), Pale Skin~(PS), Smiling~(S), Sideburns~(Sb), Straight Hair~(SH), Wearing Lipstick~(WL), Wavy Hair~(WaH), Wearing Hat~(WH), Young~(Y). As expected, the attributes of the output images aligned with the attributes of the images in the same row. For instance, the hairstyle, gender, and glasses of the real images were transferred to the output images, while the background and color characteristics of the generated images were preserved in the outputs. In fact, the average matching accuracy across all attributes is higher than $85\%$. This demonstrates the effectiveness of the proposed framework in controlling the content of the generated images while retaining other attributes influenced by the injected random vector.

\subsection{Extensive Analysis}\label{sec:analysis} 
In addition to the main experiments, we conducted in-depth analyses focusing on key aspects: the impact of the number of EM blocks on performance, the tradeoff between FID and precision, and the influence of frequency bands on image generation.

\subsubsection{Impact of EM blocks}\label{sec:extensive_emblocks}
\begin{figure}[t]
    \centering
    \includegraphics[width=0.5\textwidth]{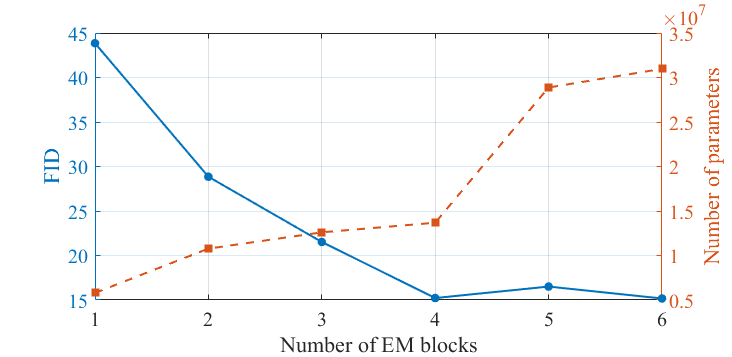} \vspace{-0.25cm}
    \caption{FID and model complexity over various numbers of EM blocks. \vspace{0.3cm}} 
    \label{fig:emblocks_vs_fid}
    \centering    
\end{figure}

To determine the structure of CFM, we conducted experiments to assess the impact of varying quantities of EM blocks on its performance, as shown in Figure~\ref{fig:emblocks_vs_fid}. Based on the experiment, we configured the CFM with 4 EM blocks. Configurations employing more than 4 EM blocks could decrease the FID score; however, they also showed a considerable escalation in model complexity, offering only marginal performance improvements compared to the 4-block structure. Consequently, the experimental result suggested that the CFM with 4 EM blocks is the optimal choice, as it ensures both satisfactory performance and manageable model complexity.

\subsubsection{Tradeoff between FID and Precision} \label{sec:extensive_tradeoff}
\begin{figure*}[t]
\begin{center}
\begin{tabular}{@{~}c@{~}c@{~}}
\includegraphics[width = 0.5\textwidth]{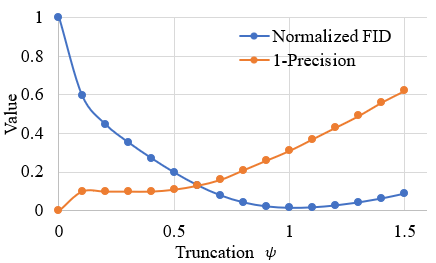} & 
\includegraphics[width = 0.5\textwidth]{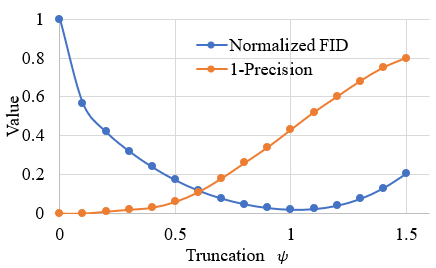}\\
{\footnotesize FFHQ} &
{\footnotesize LSUN Church} \\
\end{tabular}
\caption{Tradeoff between FID and (1-precision) over different truncation values with the maximally truncated setup~($\psi = 0$). The blue line represents the normalized FID values from 0 to 1, while the orange line represents the error~(1-precision). Note that truncation balances perceptual quality, and variation-maximizing truncation results in low diversity but high precision. As truncation is reduced, diversity increases while precision decreases.} \vspace{-0.2cm} \label{fig:tradeoff} 
\end{center}
\end{figure*}

The truncation trick~\citep{bigGANs,stylegans,truncTrick} refers to the practice of restricting the range of random noise input to enhance the similarity between generated images and real images at the expense of reducing diversity. This tradeoff between similarity and diversity is commonly observed in GANs training. When the truncation is maximized, \ie $\psi=0$, the variation in the generated images decreases due to the reduced influence of random noise. Conversely, as the truncation threshold increases, the diversity increases, but the perceptual quality of the generated images improves, making them more similar to real images~\citep{pr}. This is clearly demonstrated by the analysis results in Figure~\ref{fig:tradeoff}. In the figure, we plotted FID and (1-Precision) instead of Precision to clearly illustrate the tradeoff relationship between them. This choice is made because better performance is represented by a lower FID value, while higher values indicate better performance for Precision. 

\begin{figure*}
\begin{center}
    \begin{minipage}{0.495\linewidth}
    \centering
        \includegraphics[width= 1.0\textwidth]{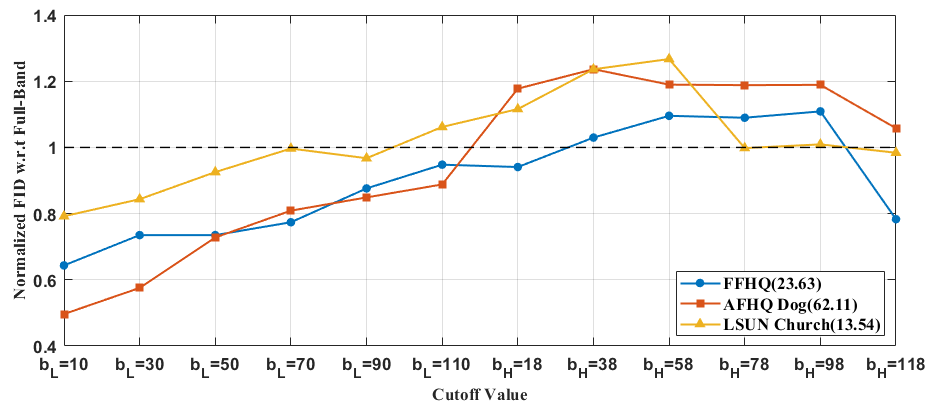}  \\
        {\footnotesize (a)}
    \end{minipage}
    \begin{minipage}{0.495\linewidth}
        \centering
        \includegraphics[width = 1.0\textwidth]{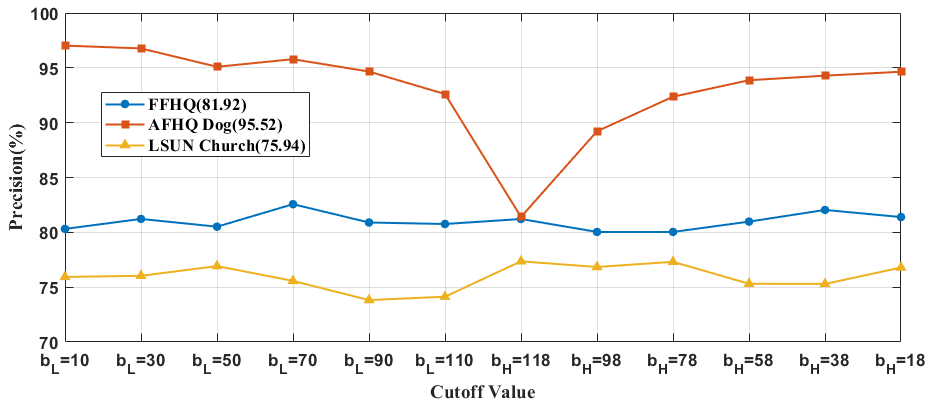}  \\
        {\footnotesize (b)}
    \end{minipage}
     \caption{Performance comparison over different cutoff frequencies in terms of FID and Precision. (a) FID of filtered images with respect to the images without filtering. (b) Precision using different cutoff values. The cutoff values of low-pass filtered images denote as $b_{\scriptscriptstyle L}$, and the values of high-pass filtered images denote as $b_{\scriptscriptstyle H}$. The values next to the dataset names represent the FID and Precision of unfiltered images.} \vspace{-0.1cm}\label{fig:performance_with_different_frequencies}
\end{center}
\end{figure*}

The observed tradeoff relationship can be directly applied to interpret the performance of the proposed framework, as reported in Section~\ref{sec:similarity_results}, in the context of its objective. The proposed framework aims to generate output images that closely resemble user-selected, specific real-world images. As a result, the precision of the generated images is significantly improved, indicating a higher percentage of accurate matches with the reference images. However, this emphasis on content preservation might come at the expense of overall statistical diversity, leading to higher FID scores compared to other methods. The intentional tradeoff aligns with the primary objective of the framework, which is to afford users control over the content of the generated images while maintaining an acceptable level of quality, as observed in the examples in Figures~\ref{fig:ffhq_results}, \ref{fig:afhq_results}, and \ref{fig:lsun_results}. This interpretation is supported by the experimental results presented in Table~\ref{tab:evaluation} and Figure~\ref{fig:performance_with_different_frequencies}. 

\subsubsection{Frequency Analysis}\label{sec:extensive_frequency}
In order to analyze the influence of frequency components on image generation, we experimented by generating images using different cutoff frequencies ($b_{\scriptscriptstyle L}$ and $b_{\scriptscriptstyle H}$) with (\ref{eq:lowpass}) or (\ref{eq:highpass}), along with (\ref{eq:dft}) and (\ref{eq:idft}). We applied this to input images with different frequency limits, enabling us to examine the correlation between frequency manipulation and the resulting generated images. 

Figure~\ref{fig:performance_with_different_frequencies}(a) shows the ratio of FID corresponding to different bandwidths of low-pass filtering using (\ref{eq:lowpass}) and high-pass filtering employing (\ref{eq:highpass}) with respect to the FID of images without filtering. The values next to the datasets' names represent the FID of the unfiltered images for reference purposes. Overall, encoding frequency-selected images into feature embeddings tended to result in improved FID scores. Interestingly, the FID scores improved as more high-frequency components were removed through lowpass filtering. On the other hand, maintaining only high-frequency components did not have a significant impact on the improvement of FID scores, although it yielded better results compared to full-band images. Likewise, increasing the bandwidth of high-pass filtered images led to better FID scores. Based on these observations, we conclude that restricting certain frequency bands improves the quality of generated images in image generation. In particular, maintaining a narrow band of low-frequency components tends to produce higher-quality images. This could be attributed to the characteristics of neural networks, which may face challenges in learning high-frequency components during training.

\begin{figure}[t]
    \centering
    \includegraphics[width=0.6\textwidth]{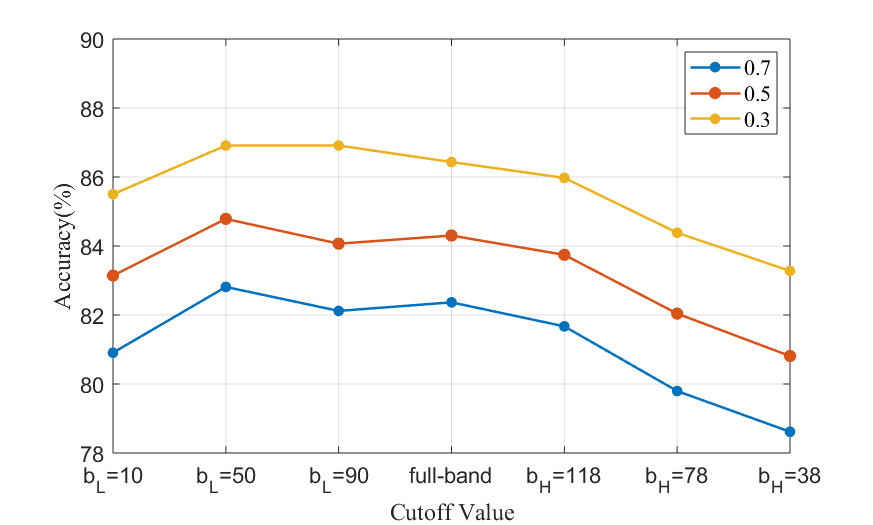} \vspace{-0.25cm}
    \caption{Average accuracy of $40$ attribute classifiers with various cutoff values and classification thresholds.} 
    \label{fig:attributes_acc}
    \centering    
\end{figure}

As already pointed out in Section~\ref{sec:experimental_results}, the tradeoff relationship between FID scores and Precision can also be observed in the graphs in Figure~\ref{fig:performance_with_different_frequencies}(a) and (b). This relationship was particularly evident in the models utilizing low-passed images, as depicted in Figure~\ref{fig:performance_with_different_frequencies}(b). Specifically, as Precision increased, the FID score also increased, indicating a decline in the similarity of the overall distribution of real data. Although the tradeoff in high-pass filtered images was less clear, there were instances where higher Precision corresponded to higher FID scores.

\begin{figure*}[t]
    \centering
    \begin{tabular}{@{}c@{~}c@{~}c@{~}c@{~}c@{~}c@{~}c@{~}c@{~}c@{~}c@{~}}
        \includegraphics[width = 0.09\textwidth]{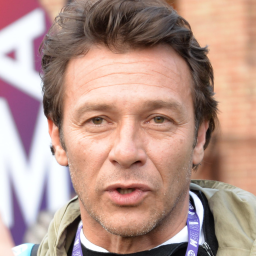} & &
        \includegraphics[width = 0.09\textwidth]{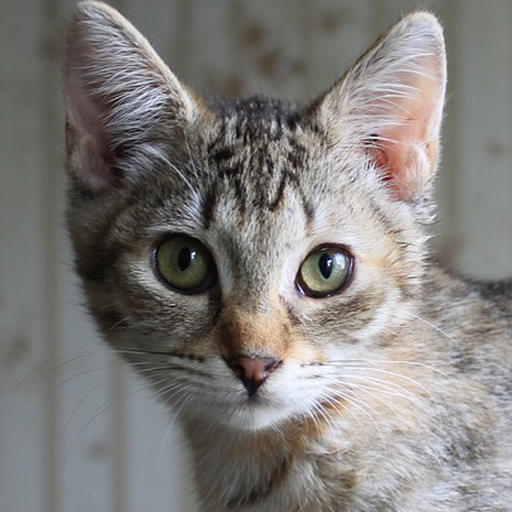} & &
        \includegraphics[width = 0.09\textwidth]{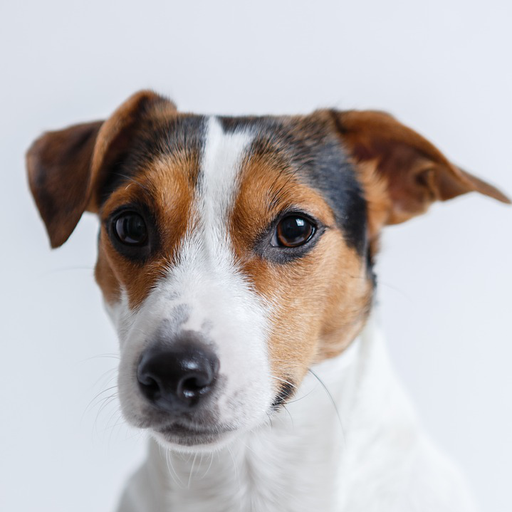} & &
        \includegraphics[width = 0.09\textwidth]{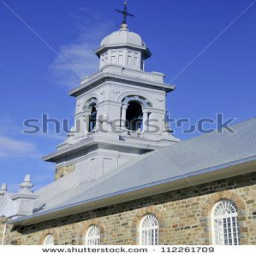} & &
        \includegraphics[width = 0.09\textwidth]{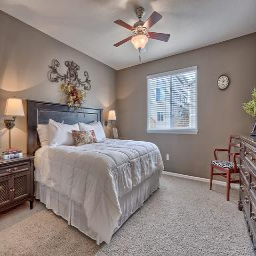} & \\
        
        \includegraphics[width = 0.09\textwidth]{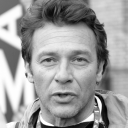} &
        \includegraphics[width = 0.09\textwidth]{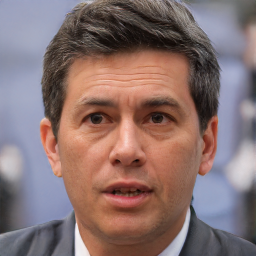} &
        \includegraphics[width = 0.09\textwidth]{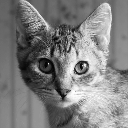} &
        \includegraphics[width = 0.09\textwidth]{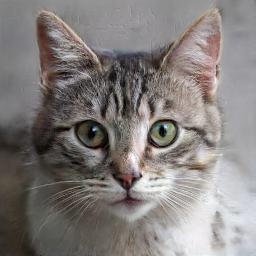} &
        \includegraphics[width = 0.09\textwidth]{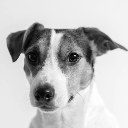} &
        \includegraphics[width = 0.09\textwidth]{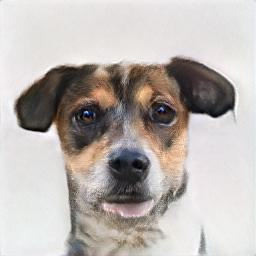} &
        \includegraphics[width = 0.09\textwidth]{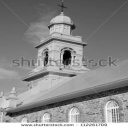} &
        \includegraphics[width = 0.09\textwidth]{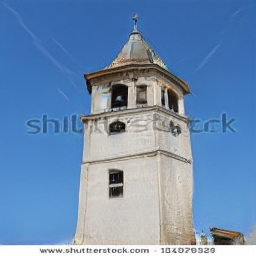} &
        \includegraphics[width = 0.09\textwidth]{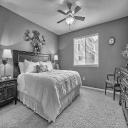} &
        \includegraphics[width = 0.09\textwidth]{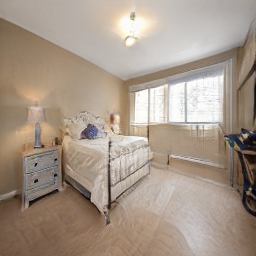} \\

        \includegraphics[width = 0.09\textwidth]{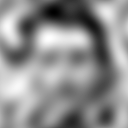} &
        \includegraphics[width = 0.09\textwidth]{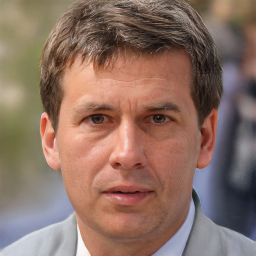} &
        \includegraphics[width = 0.09\textwidth]{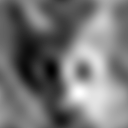} &
        \includegraphics[width = 0.09\textwidth]{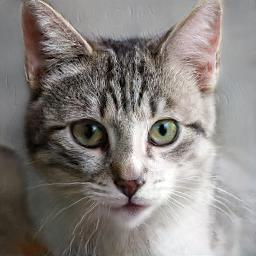} &
        \includegraphics[width = 0.09\textwidth]{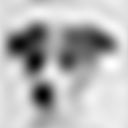} &
        \includegraphics[width = 0.09\textwidth]{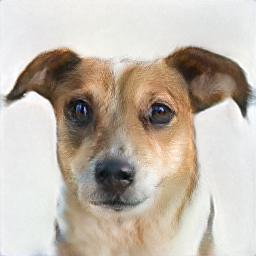} &
        \includegraphics[width = 0.09\textwidth]{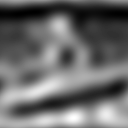} &
        \includegraphics[width = 0.09\textwidth]{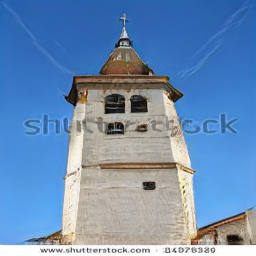} &
        \includegraphics[width = 0.09\textwidth]{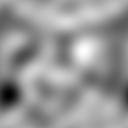} &
        \includegraphics[width = 0.09\textwidth]{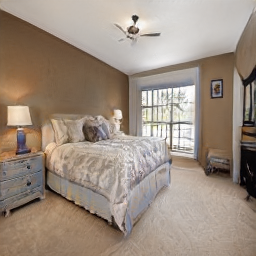} \\

        \includegraphics[width = 0.09\textwidth]{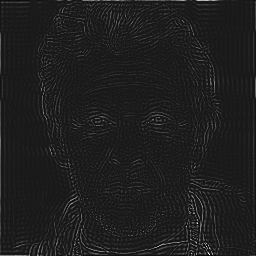} &
        \includegraphics[width = 0.09\textwidth]{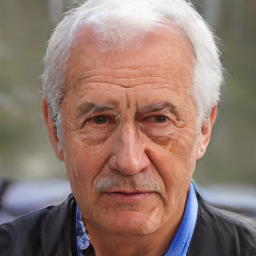} &
        \includegraphics[width = 0.09\textwidth]{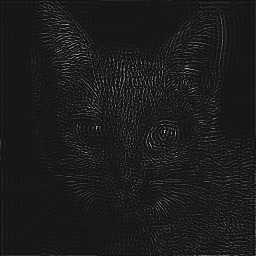} &
        \includegraphics[width = 0.09\textwidth]{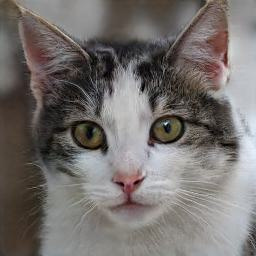} &
        \includegraphics[width = 0.09\textwidth]{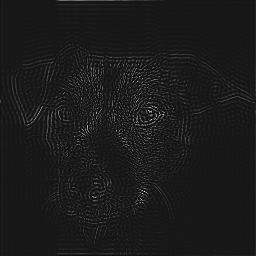} &
        \includegraphics[width = 0.09\textwidth]{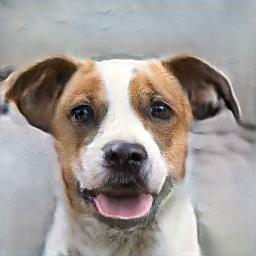} &
        \includegraphics[width = 0.09\textwidth]{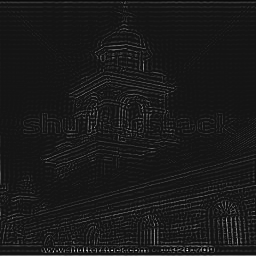} &
        \includegraphics[width = 0.09\textwidth]{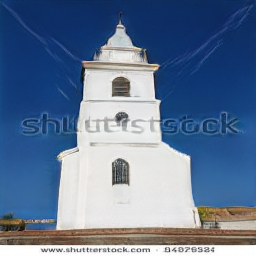} &
        \includegraphics[width = 0.09\textwidth]{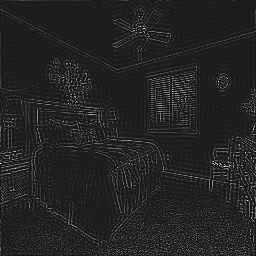} &
        \includegraphics[width = 0.09\textwidth]{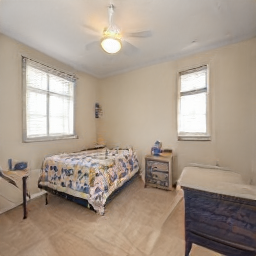} \\

        \multicolumn{2}{c}{\footnotesize FFHQ} &
        \multicolumn{2}{c}{\footnotesize AFHQ Cat} &
        \multicolumn{2}{c}{\footnotesize AFHQ Dog} &
        \multicolumn{2}{c}{\footnotesize LSUN Church} &
        \multicolumn{2}{c}{\footnotesize LSUN Bedroom} \\
    \end{tabular}

  \caption{Examples of the generated image with different frequency bands. The first row displays the real-world images. On the left side of each dataset, the frequency-filtered versions with different bandwidths are displayed, and on the right side are the output images. The second row exhibits results using full-band images. The third row displays low-passed images with $b_{\scriptscriptstyle L}=30$. Lastly, the fourth row shows high-passed images with $b_{\scriptscriptstyle H}=110$. For better visualization, high-pass components larger than the average of the high-pass images are emphasized by multiplying a scalar while suppressing other values.} \label{fig:images_with_different_cutoff}
\end{figure*}

In addition, Figure~\ref{fig:attributes_acc} displays the attribute classification accuracy on the FFHQ dataset for various cutoff values and classification thresholds~(0.3, 0.5, 0.7). To examine the influence of frequency bands on content preservation, we conducted experiments similar to those described in Section~\ref{sec:content_results}, but using different frequency-selected images. The attribute classification results indicate that wider bandwidths of low-pass filters tended to preserve the content characteristics to a slightly greater extent. This observation is consistent with the analysis of Precision, where increasing precision with wider low-pass bandwidth allowed for the preservation of more attributes. Conversely, narrow bandwidth high-pass filters showed a slight advantage in transferring content attributes. While the performance differences may not be significant, the experimental results suggest that manipulating low-frequency components, which contain more abundant spatial information, can assist in controlling the content of generated images. This provides valuable insight into content manipulation techniques. 

We also visualize how the output images look like with different cutoff frequencies in Figure~\ref{fig:images_with_different_cutoff}. These examples provide qualitative evidence that retaining low-frequency components was more effective in aligning the content with the guiding images compared to keeping high-frequency components. This reaffirms the fact that preserving only low-frequency components can enhance image quality, as indicated by the FID scores, and result in higher Precision, although the impact on Precision is minimal, as observed in the graphs of Figure~\ref{fig:performance_with_different_frequencies}. 

\section{Conclusion}\label{sec:conclusion}
In this paper, we introduced a novel encoder architecture designed to generate images that can retain the content of real-world images, offering users control over image generation. To assess the effectiveness of the proposed framework in controlling image content, we conducted attribute classification and compared characteristics of real-world and generated images. The experimental results demonstrated that the proposed framework successfully transferred content from real-world images to the generated ones, preserving an average of 85\% of the input attributes from real face images in the synthetic images of the FFHQ dataset. Furthermore, we investigated the impact of frequency bands on image generation. The associated findings revealed that low-frequency components had a more preferable effect on the FID score, while high-frequency components introduced noise and increased the FID score. We believe this study will inspire further advancements in desired content generation for image generation applications.

\bibliographystyle{elsarticle-harv}
\bibliography{refs}

\end{document}